\documentclass[preprint,11pt,3p,authoryear]{elsarticle}
\usepackage{timet,color}
\usepackage[urlcolor=blue,citecolor=black,linkcolor=black,hidelinks]{hyperref}
\usepackage{amsmath}
\usepackage{amssymb}
\usepackage{array}
\usepackage{xcolor,colortbl}
\definecolor{mygray}{gray}{0.8}
\usepackage{multirow}
\usepackage{booktabs}
\usepackage{graphicx}
\graphicspath{ {./images/} }
\usepackage[capitalise]{cleveref}
\usepackage{setspace}
\usepackage{xcolor}
\usepackage{soulutf8}
\renewcommand\hl[1]{#1}
\journal{Computers and Geotechnics}

\begin{document}

\begin{frontmatter}

\title{Using explainability to design physics-aware CNNs for solving subsurface inverse problems }

\author[1]{J. Crocker *$^,$}
\author[1]{K. Kumar}
\author[2]{B. Cox}
\affiliation[1]{The University of Texas at Austin, Department of Civil, Architectural, and Environmental Engineering, Austin, TX, USA, 78712}
\affiliation[2]{Utah State University, Department of Civil and Environmental Engineering, Logan, UT, USA, 84322}

\label{firstpage}

%\maketitle

\begin{abstract}
\doublespacing
We present a novel method of using explainability techniques to design physics-aware convolutional neural networks (CNNs). We demonstrate our approach by developing a CNN for solving an inverse problem for shallow subsurface imaging. Although CNNs have gained popularity, the development of CNNs remains an art, as there are no clear guidelines regarding the selection of hyperparameters that provide the best network. While optimization algorithms can select hyperparameters automatically, these methods develop networks with high predictive accuracy while disregarding model explainability (descriptive accuracy). The field of Explainable Artificial Intelligence (XAI) addresses the absence of model explainability by providing tools to evaluate the internal logic of networks. In this study, we use the explainability methods Score-CAM and Deep SHAP to select hyperparameters (e.g., kernel size and network depth) to develop a physics-aware CNN for shallow subsurface imaging. We begin with an Encoder-Decoder network, which uses surface wave dispersion images to generate 2D shear wave velocity images. Through model explanations, we find that a shallow CNN using two convolutional layers with an atypical kernel size of 3$\times$1 yields comparable predictive accuracy but increased descriptive accuracy. We believe this method can be used to develop networks with high predictive accuracy while providing inherent explainability.
   
\end{abstract}

\end{frontmatter}
\doublespacing
\section{Introduction}
\begin{figure}[hbt!]
 \includegraphics[width = \linewidth]{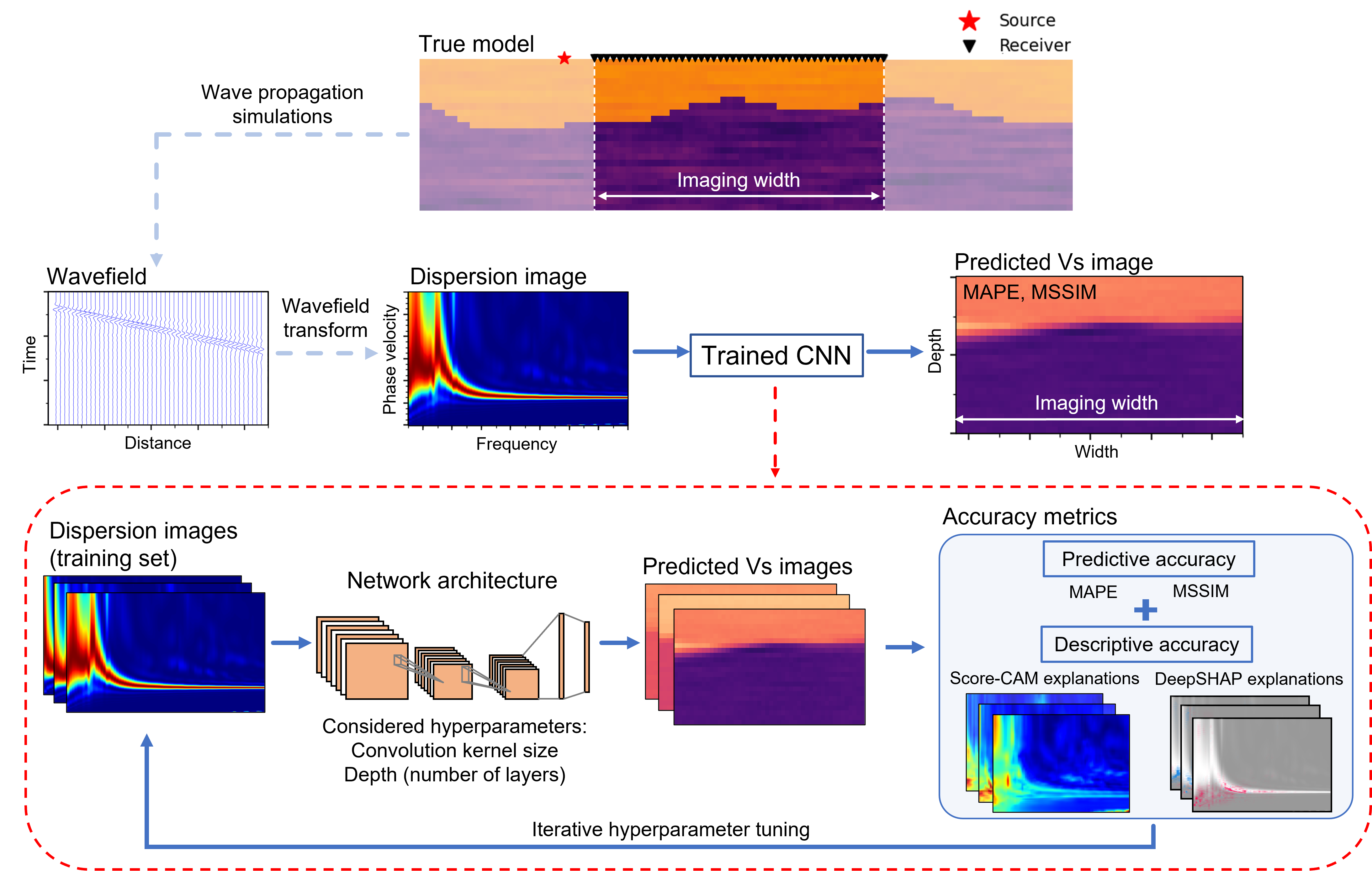}
 \vspace{-.75cm}
   \caption{The proposed framework for developing a convolution neural network (CNN) for solving an inverse problem for shallow subsurface imaging. The CNN takes surface wave dispersion images as inputs to generate 2D shear wave velocity images. The development process begins with the training of a proposed CNN. The network’s predictive and descriptive accuracies are evaluated and hyperparameter tuning is performed accordingly until a CNN is found that yields logical predictions while maintaining a high predictive accuracy.}
   \label{fig:overview}
\end{figure}

In recent years, non-invasive techniques like surface wave testing have become increasingly popular for subsurface imaging due to their computational efficiency and ease of field data acquisition. Surface wave techniques involve collecting experimental wavefields recorded at the ground surface and solving an inverse problem to identify a subsurface profile that matches the characteristics of the recorded wavefield measurements. The inverse problem involves iteratively modifying a 1D subsurface model and solving a forward wave propagation problem (e.g., the Thomson-Haskell transfer matrix approach;~\citealp{has,thom}) until a subsurface model that yields theoretical data comparable to the experimental field data is found. This inverse problem, referred to as surface wave inversion, is complex and challenging, as it is non-unique and ill-posed~\citep{ct,foti,fo,vc}. Traditional surface wave inversions also lose the true 2D/3D nature of the subsurface contained in the recorded waveforms by only matching against theoretical 1D subsurface models. Although full-waveform inversion (FWI) can provide true 2D/3D subsurface images, it is much more time-consuming in terms of both field data acquisition and inversion compared to traditional surface wave methods. Hence, there is a need for alternative subsurface imaging methods to surface wave testing and FWI that are computationally efficient and accurate. 
A promising alternative to traditional surface wave inversion for developing subsurface images is using deep neural networks, such as convolutional neural networks (CNNs). An appropriately trained CNN can quickly and accurately generate subsurface images directly from experimental data without needing to solve the forward problem. Although the adoption of deep learning for seismic imaging is relatively recent \citep{dr}, many authors have successfully used CNNs to image the deep subsurface~\citep{li,sun,wang,wul,zhe}. However, limited research has been conducted on using CNNs for shallow subsurface imaging (depths less than ~30 m). Shallow subsurface imaging is particularly challenging because material properties vary significantly over short distances (vertically and laterally) and the various components of the elastic wavefield (i.e., compression, shear, and surface waves) are mixed, having not yet propagated far enough to spread out from one another. \citet{vant} developed a CNN for shallow subsurface imaging of soil-over-rock profiles that accepts active-source seismic wavefields recorded by a linear array of receivers as input to rapidly generate 2D subsurface images of shear wave velocity (Vs). \citet{abb} expanded on this network by demonstrating the possibility of generalizing such CNNs to a variety of field-testing configurations using a frequency-domain approach rather than a time-domain approach. The input of the frequency-domain CNN is dispersion images calculated from active-source wavefields recorded by a linear array of receivers placed on the ground surface, exactly like those collected from Multi-channel Analysis of Surface Waves (MASW) testing, as illustrated in the top half of~\cref{fig:overview}. While their CNN was trained on synthetic data, it yields a high predictive accuracy on field data collected at a site with corroborating subsurface ground truth. Despite the promising results, CNNs are traditionally designed using a standard trial-and-error approach by varying the network architecture and performing hyperparameter tuning until the network yields a high predictive accuracy. This blind trial-and-error approach ignores the logic, or explanations, behind model predictions, limiting their trustworthiness and applicability. In this work, we propose a framework for designing a physics-aware neural network architecture using model explainability to select the network’s hyperparameters, leading to a network that has high predictive accuracy and inherent explainability.

Deep learning models vary in accuracy, complexity, and interpretability. Simple models, such as linear regression, classification rules, and decision trees, use linear functions that are highly interpretable, but lack predictive accuracy. Highly non-smooth, non-linear, or multilayer models, such as CNNs, trade this interpretability for high accuracy. CNNs are multilayer neural network models capable of solving complex problems such as image recognition, but offer no explanations regarding their internal logic or decision making process (i.e., they are black box in nature). Hence, it is difficult to assess the physics behind their predictions. Typically, models are designed to have high predictive accuracy (i.e., the accuracy of the CNN’s predictions versus known data) but low descriptive accuracy (i.e., how accurately the model explains why it made that prediction; \citealp{mur}). Therefore, model explainers are required to address the low descriptive accuracy of these black box models. An alternative to training black box models is to incorporate a physics-informed loss function. Although Physics Informed Neural Networks (PINNs) are trained to conserve physical properties, they do not guarantee a physically-consistent response, and the reasoning behind their predictions still remains a black box \citep{rai}. The field of Explainable Artificial Intelligence (XAI) offers tools and techniques to comprehend the results of deep learning models. These tools include model explainers such as LIME \citep{rib}, class activation maps (CAMs; \citealp{zhou}), eXplanation with Ranked Area Integrals (XRAI; \citealp{kap}), and integrated gradients (IG; \citealp{sund}), which provide post hoc explanations of model predictions without the need for retraining. These post hoc explanations interpret the model’s decision-making process after its development, thus improving its trust without compromising the model’s predictive accuracy \citep{bie,lau,lou,ra,raj,mit,zech}. However, the post hoc nature of these explainers also means the explanations are not rigorous and are fallible, and may only capture a specific aspect of a model \citep{lin}. Moreover, model explanations are limited mainly to understanding and interpreting predictions rather than designing and optimizing the deep learning model, and thus are disregarded during the design of neural networks.

Despite advances in neural network algorithms, the process of designing a neural network remains an art. Developing a CNN requires careful tuning of more than 100,000 hyperparameters, such as the kernel size, depth of the network, type of layers, learning rate, or optimization algorithm, through a trial-and-error iterative process of minimizing its predictive error. Without a clear set of guidelines for selecting hyperparameters, model tuning uses a brute-force approach to optimize the network until it yields a high predictive accuracy without considering its descriptive accuracy. Approaches such as Bayesian optimization or search algorithms offer the automatic selection of hyperparameters in CNNs that yield a low error; however, these approaches also disregard model explanations \citep{hinz,lee,lh,sno,wu}. The design of CNNs by minimizing the predictive error does not necessarily yield models that are logically consistent. The network may not capture the physics or explain the logic behind its predictions, potentially yielding untrustworthy results. Although descriptive accuracy metrics may offer post hoc justification of model behavior, it is often considered separately from the model’s predictive accuracy, leading to a disconnect between model design and use. This disconnect is particularly noticeable in the field of geophysics, as most models are designed only using predictive accuracy while offering no explanations \citep{abb,ara,ck,hu,li,lu,vant,yang,zhu}.

In this work, we address the disconnect between predictive and descriptive accuracy by developing a framework for creating physics-aware neural networks by integrating model explainability into the network development process. \hl{Rather than using a physical loss function, we create a "physics-aware" network by guiding it towards learning the physical properties associated with an inverse problem.} We create a CNN for solving an inverse problem associated with shallow 2D subsurface imaging, where model explainability is used to iteratively select network hyperparameters, such as the depth of the network and the size of the convolutional layers. An overview of our approach is shown in~\cref{fig:overview}. The CNN is trained on 80,000 synthetic subsurface images representing the common geological condition of soil with varying thickness and stiffness overlying undulating rock of varying stiffness. We use finite difference elastic wave propagation simulations and MASW-style wavefield transformation methods to obtain surface wave dispersion images that capture key features of each subsurface model. These key features of the subsurface model are expected to be best predicted by the high-power regions (i.e., red colors) in the dispersion images, which represent the fundamental and potentially higher-modes of surface wave propagation. Thus, our dataset consists of pairs of known 2D subsurface velocity images and their corresponding surface wave dispersion images. \hl{We choose to train the network using only dispersion images, as the high-power modal trend (discussed in detail in a later section) is relatively stable for a given subsurface model, regardless of the testing configuration used to sample the model.} The network is first trained with respect to its predictive accuracy, but we include an additional step that evaluates the descriptive accuracy of the trained network on the validation dataset. By doing so, we can interpret the CNN’s decision-making and determine if its predictions are logical. If we determine the network makes illogical predictions, the development process is revisited, and various hyperparameters are iteratively tuned to improve the network. 

In this work, we propose a framework for performing hyperparameter tuning and developing a neural network architecture that improves its descriptive accuracy while maintaining a high predictive accuracy. The aim of this work is not to generalize the network to geologic conditions outside the bounds of the training dataset or to improve the accuracy of its predicted 2D shear wave velocity profiles; we instead guide the network to make logical decisions based on the current problem. We believe that progress made in this area will benefit the development of CNNs for solving inverse problems related to subsurface imaging.

\subsection{Convolutional neural networks for solving an inverse problem for shallow subsurface imaging}

Convolutional neural networks are a type of deep learning model that excels at image classification and generation. CNNs are typically composed of one or more convolutional layers, each followed by a pooling layer, and commonly ending with a fully connected layer. Convolutional layers consist of a set of filters defined by a kernel size. These filters are used to perform convolution on the input, and an activation function is applied to obtain feature maps, which contain information about the spatial features of the input. Pooling layers are used to reduce the dimensionality of each feature map representation, thus reducing the number of learnable parameters and computational costs. Finally, after multiple cycles of convolution and pooling layers, fully connected layers may be used to learn non-linear combinations of the feature maps. 

\begin{figure}[hbt!]
 \includegraphics[width=\linewidth]{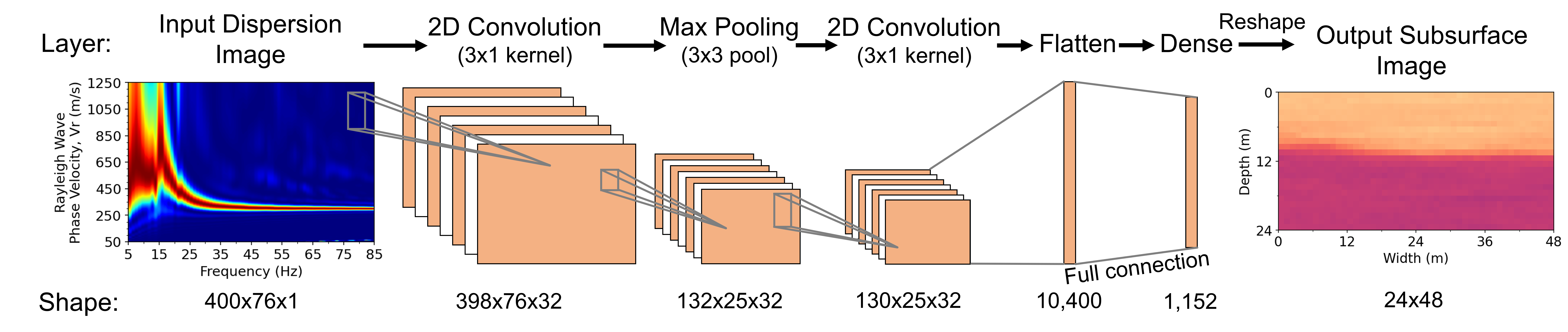}
 \vspace{-0.7cm}
   \caption{The architecture of a convolutional neural network used to generate subsurface images using MASW-style data. The structure, kernel size of the convolutional layers, and depth are selected considering the network’s predictive and descriptive accuracy.}
   \label{fig:cnnarch}
\end{figure}

With the addition of each layer, CNNs improve their predictive accuracy as each additional layer extracts more complex information from the input image. For example, the first layer typically identifies the layer boundaries within an image, and subsequent layers identify complex features such as gradients and curvatures. Many different metrics, such as mean square error (MSE), mean absolute error (MAE), or mean absolute percentage error (MAPE), are used to evaluate the accuracy of the network’s predictions, and typically networks are developed to have high predictive accuracy. An example of a CNN architecture is provided as~\cref{fig:cnnarch}, which is the final network developed in this study, as described in later sections. Note that we use only two 2D convolutional layers separated by a single max pooling layer ending with a final fully connected layer. For more information on CNNs, please refer to \citet{good}.

\section{Accuracy metrics}
\subsection{Predictive accuracy metrics}

In this study, we define the global accuracy of our network’s predictions based on the entire subsurface image rather than a specific feature (e.g., only the Vs of the overlying soil layer or depth to the rock layer). Therefore, we measure the accuracy of our networks across 20,000 testing image pairs using two global error metrics: (1) MAPE and (2) mean structural similarity index (MSSIM; \citealp{wb}). MAPE measures the mean of the absolute value of the pixel-by-pixel percent error of each predicted Vs image compared to the true image. We use MAPE to relate to the physical units of each image (i.e., Vs magnitude), while other common metrics such as MSE or MAE only correspond to normalized images. A limitation of MAPE is that it does not consider the structure of an image. For example, if a true image and its predicted counterpart are reordered pixel-by-pixel in the same way, then the MAPE does not change despite the changes in images. Therefore, we also consider the SSIM index. The SSIM index compares a predicted image to its true counterpart using three criteria: luminance, contrast, and structure. This index is computed on a pixel-by-pixel basis to create an SSIM index map, and a mean value may be taken as the MSSIM to provide a single overall measure of image similarity for each prediction. The MSSIM ranges from zero for predictions with poor similarity to their respective true images to one for a prediction image that is identical to the true image. In this study, we follow the parameters recommended by \citet{wb} to calculate MSSIM with a dynamic range equal to the difference between the maximum and minimum Vs values of the true subsurface profiles in the testing dataset.

\begin{figure}[hbt!]
 \includegraphics[width=0.9\linewidth]{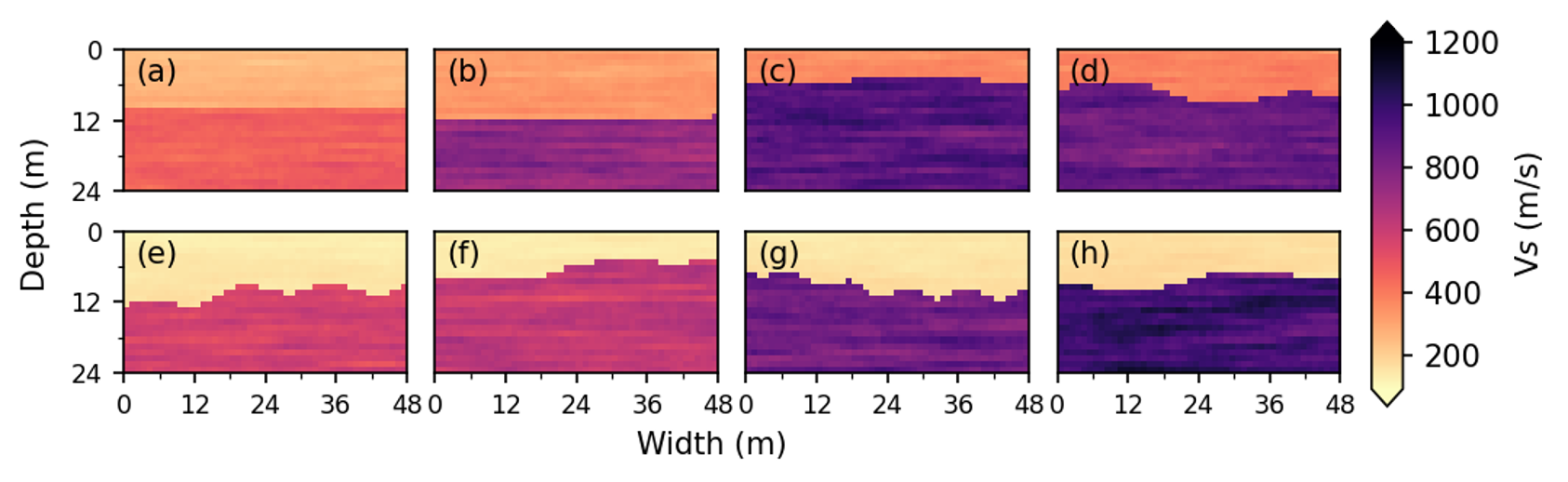}
 \centering
 \vspace{-0.5cm}
   \caption{A selected set of true synthetic subsurface soil-over-rock Vs images ranging from simple images (a-d) to complex images (e-h).}
   \label{fig:true_models}
 \includegraphics[width=0.9\linewidth]{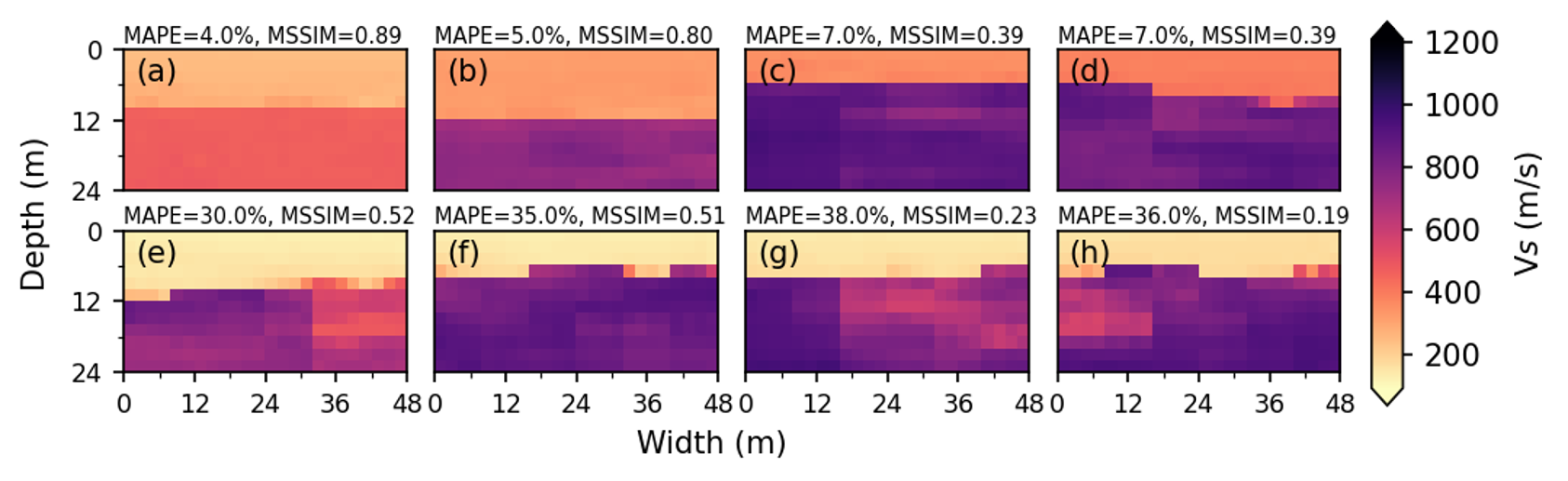}
 \vspace{-0.5cm}
   \caption{The 
   A selected network’s predictions of the subsurface images depicted in~\cref{fig:true_models}.}
   \label{fig:boundaries}
\end{figure}

We choose to use both MAPE and the MSSIM index to provide two perspectives regarding the predictive accuracy of our networks. As MAPE provides an error in physical units, we propose that this measures the overall accuracy of the predicted Vs values in each subsurface image despite not considering the overall image structure. In contrast, the MSSIM index considers the predicted Vs values in combination with the image structure, and therefore we hypothesize that it reflects the accuracy of the predicted layer boundary in each subsurface image. We illustrate the use of both metrics using a subset of eight true subsurface Vs images from the testing dataset (\cref{fig:true_models}), and the respective predictions of a \hl{selected CNN}, as described in a later section, with MAPE and MSSIM values (\cref{fig:boundaries}). 

Note that the images shown in~\cref{fig:true_models} and~\cref{fig:boundaries} were specifically selected to better illustrate the following trends, which are consistent for all the images analyzed throughout this study. Comparing Figs.~\labelcref{fig:boundaries}a and b to the true images in Figs.~\labelcref{fig:true_models}a and b, we see the predictions are visually similar, which is reflected by their low MAPE values and high MSSIM indices. Specifically, the Vs values and the layer boundaries, which are simply the depths to the rock layer, are predicted well in these images. In Figs.~\labelcref{fig:boundaries}c and d, we see the Vs values are generally predicted well (relatively low MAPEs), while the undulations of the layer boundary are relatively smoothed (relatively low MSSIM indices that indicate structural differences at the interface). In Figs.~\labelcref{fig:boundaries}e and f, the predicted Vs values are higher than the true images (high MAPEs), and the layer boundaries are overly smoothed, although the average depths to the rock layer are captured (relatively low MSSIM indices). Finally, in Figs.~\labelcref{fig:boundaries}g and h, we see the predicted Vs values are generally lower than the true images (high MAPEs) and undulations in the layer boundary are lost (\cref{fig:boundaries}g) or inverted (\cref{fig:boundaries}h; low MSSIM indices). From these results, we conclude that the magnitude of MAPE is strongly affected by the overall predicted Vs values, while the magnitude of the MSSIM index strongly correlates with the prediction of layer boundaries, and thus we use both metrics to evaluate the predictive accuracy of our networks. Note that the CNN is trained only using the mean absolute error (MAE) loss function, as described later.

\subsection{Descriptive accuracy metrics}

We use post hoc explainability methods to comprehend the black box CNN model used in this study. As stated previously, there is no single post hoc explainability method that can cover all the complexities of a CNN, as they are not rigorous. Therefore, we use two different descriptive accuracy metrics to review the internal logic of our CNNs: Score-weighted Class Activation Mapping (Score-CAM; \citealp{wa}) and Deep SHAP~\citep{ll}. We present a brief overview of these methods, but the reader is encouraged to review the literature for a more in-depth discussion of the algorithms and implementations.

\subsubsection{Score-weighted Class Activation Mapping (Score-CAM)}

Heatmaps are often used as a model explainer, as they provide a visual representation of a network’s internal logic. One popular method is the Gradient-weighted Class Activation Mapping (Grad-CAM; \citealp{sel}) algorithm, which is a post hoc explainer that uses the gradients of a class, or the classification of an image, flowing into a CNN’s final convolutional layer to create heatmaps associated with that class. It is a generalization of the method used to create class activation maps (CAMs) by \citet{zhou}, as it does not require using a particular architecture (i.e., a final convolutional layer followed by a global average pooling layer and fully connected layer) to develop the heatmaps

Although a popular explainability method, recent work shows that Grad-CAM may generate heatmaps that highlight irrelevant regions in the input image that are not used by a network for prediction \citep{chat,wa}. This is caused by the gradient averaging step, as the computed weights used to create a heatmap do not adequately capture each feature map's importance. For example, two feature maps may have similar weights, despite one feature map having a lower contribution to the output image, as demonstrated by \citet{wa}. Hence, we use Score-CAM, proposed by \citet{wa}, which uses score-based weights to obtain a final heatmap rather than using gradients to weight feature maps. Score-CAM extracts the feature maps ($A_{k}$) generated by the last convolutional layer in a network for a given input image, where \emph{k} is the number of feature maps. Each feature map is then treated as a mask on the original input image, and the masked image is passed through the network to obtain its score of a target class. For image generation applications, each pixel is assigned a class rather than an individual class for the whole image. The scoring process is repeated \emph{k} times to obtain the score-based weights ($\alpha_{k}^c$) and the final heatmap is generated using a linear combination of $\alpha_{k}^c$ with $A_{k}$. Score-CAM uses ReLU to preserve positive values in the heatmap while setting negative values to zero, so the heatmap only highlights parts of the input image that contribute to the output. Additionally, the authors note that while the last convolutional layer in a network is the preferable choice for use with Score-CAM, this process can be applied to any convolutional layer to obtain its associated heatmap. An example of a Score-CAM heatmap is shown in~\cref{fig:overview} under the descriptive accuracy header, while its interpretation is discussed in a later section.

\subsubsection{Deep Learning Important FeaTures (DeepLIFT) with Shapley values (Deep SHAP)}

Shapley values are used to quantify the relative contributions of various input features to an output. \citet{ll} used this concept to create SHAP (SHapley Additive exPlanation) values, which quantify the relative change in the expected model prediction caused by the introduction of each new feature to the model. Deep SHAP~\citep{ll} combines SHAP values with the Deep Learning Important FeaTures method (DeepLIFT; \citealp{shr}). The DeepLIFT method explains the difference in an output from a set “reference” output by using the differences between perturbed inputs and a set “reference” input. Specifically, the relative contribution caused by the differences between neurons for an input image and its reference input must equal the difference between an output and its reference output. These relative contributions are separated into positive and negative components, and a linear composition is used to assign contribution scores for each neuron to its immediate inputs. Although similar to other backpropagation methods, SHAP addresses the issue of saturation that is inherent in gradient-based methods. 

SHAP values provide a method of measuring additive feature importance, which allows it to be combined with other additive feature attribution methods, such as DeepLIFT. As DeepLIFT can linearize the non-linear components of a deep neural network (i.e., through its linear composition rule) and its input images are assumed to be independent of one another, DeepLIFT can approximate SHAP values, leading to Deep SHAP. In this case, SHAP values measure the relative contributions of perturbed inputs versus the unperturbed input. Like Shapley values, the magnitude of SHAP values refers to the degree of influence an input has on output, while positive and negative values correspond to positive and negative contributions to the output. We compute Deep SHAP values using the open-source package shap \citep{ll}. An example Deep SHAP result is shown in~\cref{fig:overview} under the descriptive accuracy header, while its interpretation is discussed in a later section.

\section{Dataset creation}

Our dataset contains two components: (1) a set of 2D subsurface Vs images and (2) a corresponding set of surface wave dispersion images (one dispersion image for each 2D subsurface model, explained in the following sections). We begin by developing 2D subsurface Vs images, then use these Vs images as the basis for generating corresponding compression wave velocity (Vp) and mass density images using common relationships. The Vs, Vp, and mass density images together  form the subsurface models needed for elastic wave propagation simulations. Following subsurface model development, we use 2D finite difference numerical wave propagation simulations with these subsurface models to generate waveforms recorded by a linear array of receivers at the ground surface. These waveforms are then transformed into dispersion images that capture key features of each subsurface model. In this study, only the pairs of subsurface Vs images and their corresponding dispersion images (refer to~\cref{fig:model_ex}) are considered for CNN development. We discuss additional details of the dataset development in the following sections.

\subsection{Development of subsurface Vs images}

We design each subsurface model to mimic a relatively simple but common subsurface geological condition: soil with varying thickness and stiffness overlying undulating rock of varying stiffness (i.e., soil-over-rock with an irregular interface). These subsurface models are programmatically generated using an in-house code. We create the subsurface models using a range of typical values for subsurface material characteristics, such as Vs, density, and Poisson’s ratio, as per \citet{fo}. Additionally, we also vary the undulations of the layer interface, the depth to the rock/weathered rock layer, and the depth of the groundwater table. The development process is summarized as follows: we generate a 104 m-wide and 24 m-deep subsurface Vs image using a range of typical Vs values for the subsurface materials. We separate the upper soil layer from the lower rock layer with a spatially variable and undulating interface, which is created by summing three sinusoidal waves, each controlled by a respective spatial undulation frequency and amplitude pair. Three types of layer interfaces are considered: (1) “linear”, which has nearly linear interfaces with very little to no undulations, but may be inclined (\cref{fig:model_ex}a), (2) “slightly undulating”, which has slight undulations (\cref{fig:model_ex}b), and (3) “highly undulating”, which has more frequent undulations (\cref{fig:model_ex}c). Our dataset contains 10\% linear, 60\% slightly undulating, and 30\% highly undulating subsurface models, such that the network is trained primarily on more realistic subsurface model geometries (i.e., linear and slightly undulating layer interfaces) while still accounting for more extreme or complex layer interface geometries. We place the layer interface at a variable depth between 5\textendash12 m, which completes the base subsurface Vs image. However, to create relatively more realistic subsurface images, we use random sampling to introduce lateral and vertical perturbations to each image.

We next create subsurface Vp images by relating each subsurface Vs image to Vp using a Poisson’s ratio between 0.15\textendash0.35 for the soil layers and between 0.20\textendash0.25 for the rock layers. A groundwater table is randomly placed between the surface and bottom of the Vs image, and any soil materials beneath the water table are assigned a saturated Poisson’s ratio between 0.47\textendash0.49 instead of the default range noted above. Finally, we create subsurface density images by assigning a value between 1,650\textendash2,000 kg/m$^3$ for soil and 2,100\textendash2,400 kg/m$^3$ for rock/weathered rock.
	
Following these procedures, we develop 100,000 synthetic subsurface models in total, with 80,000 for training and 20,000 for testing. Note that we only consider the subsurface Vs images in our dataset, but the Vs, Vp, and density images combined create the subsurface models on which we perform the wave propagation simulations described in the following section. To prepare the Vs images for our networks, we normalize them by the maximum Vs value in the training set. Additionally, the Vs images are trimmed to include only the 48~m beneath the center of each image so that they only represent the subsurface conditions directly beneath the linear array of receivers used to record the wave propagation simulations, as per \citet{abb}. Therefore, each input 2D Vs image is 24 m deep and 48 m wide, corresponding to an array of Vs values of size 24$\times$48$\times$1.

\subsection{Development of dispersion images}

Each dispersion image is calculated from an active-source seismic wavefield recorded by a linear array of surface sensors. We develop seismic wavefields by performing elastic wave propagation simulations with the 2D finite-difference open-source software DENISE~\citep{kohn,ko}. For brevity, we summarize our experimental configuration setup, which is similar to~\citet{abb}. The surface array used to record the wavefields consists of 48 receivers equally spaced at 1 m intervals, and the source used in our simulations was a 30 Hz Ricker wavelet, which is meant to simulate a broadband impact source like a sledgehammer or drop weight. Additionally, we use the same simulation parameters (i.e., finite difference operator, boundary conditions, etc.) and model two seconds of wave propagations through each of the 100,000 models. These simulations are performed on the Texas Advanced Computing Center’s (TACC’s) high-performance cluster Stampede2 using a single compute node with Skylake processors. 

We generate dispersion images from the seismic wavefields recorded at the surface of each model by using the frequency-domain beamforming (FDBF) wavefield transformation \citep{zy}. Additional details about the dispersion image computations may be found in \citet{abb}. The FDBF transformation was used to generate dispersion images over a common frequency range of 5\textendash80 Hz with a 1 Hz interval, resulting in 76 frequencies, and a common phase velocity range of 50\textendash1,250 m/s with a velocity interval of 3 m/s, resulting in 400 velocities. Therefore, each dispersion image corresponds to an array of phase velocity values of size 400$\times$76$\times$1. To normalize each dispersion image, we use frequency-dependent normalization, which simplifies the images for the networks compared to using absolute maximum normalization. These dispersion images are generated using the open-source Python package \emph{swprocess} \citep{van}.

\begin{figure}[!h]
 \includegraphics[width=0.8\linewidth]{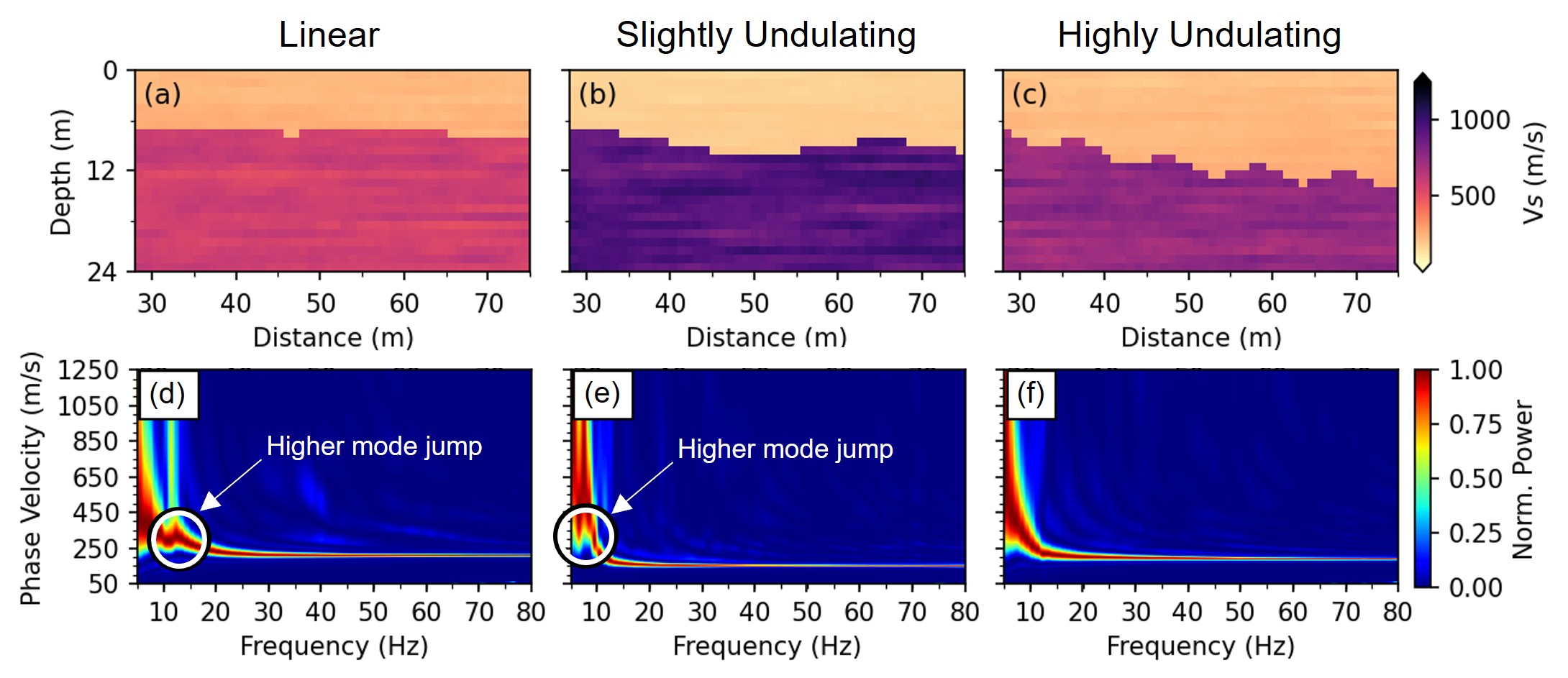}
 \centering
 \vspace{-0.5cm}
   \caption{Examples of (a-c) subsurface Vs images with linear, slightly undulating, and highly undulating layer boundaries, respectively, and (d-f) their associated dispersion input images.}
   \label{fig:model_ex}
\end{figure}

Examples of pairs of subsurface Vs images and their respective dispersion input images are shown in Figs.~\labelcref{fig:model_ex}a-c and Figs.~\labelcref{fig:model_ex}d-f, respectively. As noted above, and documented in many publications (e.g., \citealp{fo}), surface wave dispersion images capture key features of the subsurface (e.g., the velocity and layering variations). These key features are expected to be best predicted by the high-power regions (i.e., red colors) in the dispersion images, which represent the fundamental and potentially higher-modes of surface wave propagation that are excited by an active source operated on the surface. In traditional surface wave testing, it is important to discern which mode(s) are present in the dispersion image, such that appropriate, mode-specific inversions can be performed. While mode jumps from fundamental to higher-modes can be noted in some of the dispersion images (refer to Figs.~\labelcref{fig:model_ex}d and e), it is not necessary to isolate/delineate the fundamental-mode prior to network training. Rather, the entire dispersion image is used with an understanding that the network should ideally focus on learning what subsurface characteristics tend to generate the high-power trends in the dispersion image, regardless of the exact mode of surface wave propagation. Thus, for simplicity herein, we do not try to distinguish which modes are present in the dispersion images, and rather we simply refer to these fundamental and possibly higher-mode regions as the \emph{high-power modal trend}. \hl{Furthermore, we choose to use dispersion input images in this study due to the relative insensitivity of the high-power modal trend to the experimental testing configurations used to collect the wavefield and resulting dispersion data. Because the high-power modal trend is relatively stable for a given subsurface model regardless of the simulated testing configurations, a frequency-domain CNN can be generalized to different testing configurations or acquisition parameters, thus making a network trained on dispersion images a useful tool for practitioners.}

\section{Developing a convolutional neural network using explainability}

We now illustrate our proposed framework for developing a physics-aware CNN using model explainability. \hl{Note that we focus on tuning a select group of hyperparameters, such as the network's decoder structure, the kernel size of each convolution layer, and the depth of the network, as the network’s performance was most sensitive to these hyperparameters. However, the proposed framework can be used to tune any number of hyperparameters beyond those shown in this study.}

\subsection{Selecting a decoder structure}

We begin with a simple Encoder-Decoder CNN architecture, as this is one of the most widely used networks for solving inverse problems in imaging~\citep{gen}. An Encoder-Decoder CNN processes an input image $ x \in X \subset \mathbb{R}^{n} $ and maps it to a feature space $ z \in Z \subset \mathbb{R}^{d} $. The decoder produces an output $ y \in Y \subset \mathbb{R}^{l} $ from the feature map $z$. Traditionally, an Encoder-Decoder CNN is symmetric with skipped connections. We develop an asymmetric Encoder-Decoder CNN architecture to match the shape of the input dispersion images (400$\times$76$\times$1) to the shape of the output Vs subsurface images (24$\times$48$\times$1). The encoder contains five convolutional layers using a conventional 3$\times$3 kernel size, each followed by a rectified linear unit (ReLU) and a max pooling operation. 

We evaluate two commonly used decoder structures: (1) a decoder structure consisting of 2D Transpose Convolution and UpSampling layers and (2) a Dense/fully connected layer. The first trial decoder structure used in the "Transpose Convolution" network contains five 2D Transpose Convolution layers, each followed by an UpSampling layer to mirror the proposed encoder structure. We vary the kernel size for the transpose convolutional layers to match the shape of the output Vs subsurface images to the input dispersion images. The second trial decoder structure used in the "Dense" network consists of a single Dense layer containing 1152 nodes, which correspond to each pixel in our flattened Vs subsurface image (i.e., 24$\times$48$\times$1 flattened). Additionally, we choose a learning rate of 5E-4, batch size of 16, and 40 training epochs for the training of both networks. The Adam optimizer~\citep{kb} and mean absolute error (MAE) are used as the optimizer and loss function. Table~\labelcref{table:decoder} provides a detailed description of each network, and the difference between the decoder structures is highlighted. We note that validation was performed using 20\% of the training dataset, and the loss function for validation did not indicate any overfitting by the network. 

\begin{table}
\caption{Detailed descriptions of the Transpose Convolution Network (left) and the Dense Network (right). All convolution, pooling, and upsampling layers listed are 2D. Note that the only difference between the networks is the decoder structure, which is highlighted.}
\small
\label{table:decoder}
\def\arraystretch{1.2}
\vspace{0.3cm}
\begin{tabular}{>{\centering\arraybackslash}m{2cm} c >{\centering\arraybackslash}m{1.8cm} >{\centering\arraybackslash}m{1.8cm}|
>{\centering\arraybackslash}m{2cm} c >{\centering\arraybackslash}m{1.8cm} >{\centering\arraybackslash}m{1.8cm}}
\toprule
 \multicolumn{4}{c}{Transpose Convolution Network} & \multicolumn{4}{c}{Dense Network}\\
 \midrule
 Network Layer Type	& Filters & Kernel/Pool Size & Output Shape & Network Layer Type & Filters & Kernel/Pool Size & Output Shape\\
 \midrule
 Input & & & 400$\times$76$\times$1 (Dispersion Image) & Input & & & 	400$\times$76$\times$1 (Dispersion Image)\\
 Convolution & 32 & (3, 3) & 398$\times$74$\times$32 & Convolution & 32 & (3, 3) & 398$\times$74$\times$32\\
 Max Pooling & & (3, 1) & 132$\times$74$\times$32 & Max Pooling & & (3, 1) & 132$\times$74$\times$32\\
 Convolution & 32 & (3, 3) & 130$\times$72$\times$32 & Convolution & 32 & (3, 3) & 130$\times$72$\times$32\\
 Max Pooling & & (3, 1) & 43$\times$72$\times$32 & Max Pooling & & (3, 1) & 43$\times$72$\times$32\\
 Convolution & 64 & (3, 3) & 41$\times$70$\times$64 & Convolution & 64 & (3, 3) & 41$\times$70$\times$64\\
 Max Pooling & & (1, 3) & 41$\times$23$\times$64 & Max Pooling & & (1, 3) & 41$\times$23$\times$64\\
 Convolution & 128 & (3, 3) & 39$\times$21$\times$128 & Convolution & 128 & (3, 3) & 39$\times$21$\times$128\\
 Max Pooling & & (3, 3) & 13$\times$7$\times$128 & Max Pooling & & (3, 3) & 13$\times$7$\times$128\\
 Convolution & 128 & (3, 3) & 11$\times$5$\times$128 & Convolution & 128 & (3, 3) & 11$\times$5$\times$128\\
 \cellcolor{mygray}Transpose Convolution & 128 & (1, 1) & 11$\times$5$\times$128 \\
 \cellcolor{mygray}UpSampling & & (1, 2) & 11$\times$10$\times$128 \\
 \cellcolor{mygray}Transpose Convolution & 64 & (1, 2) & 11$\times$11$\times$64 \\	
 \cellcolor{mygray}UpSampling & & (1, 2) & 11$\times$22$\times$64 & \cellcolor{mygray}Flatten & & & 7040 \\
 \cellcolor{mygray}Transpose Convolution & 32 & (1, 2) & 11$\times$23$\times$32 & \cellcolor{mygray}Dense & & & 1152 (Flattened Vs Image) \\
 \cellcolor{mygray}UpSampling & & (1, 1) & 11$\times$23$\times$32 \\	
 \cellcolor{mygray}Transpose Convolution & 32 & (2, 2) & 12$\times$24$\times$32 \\
 \cellcolor{mygray}UpSampling & & (2, 2) & 24$\times$48$\times$32 \\
 \cellcolor{mygray}Convolution & 1 & (1, 1) & 24$\times$48$\times$1 (Vs Image)		\\
 \bottomrule
\end{tabular}
\end{table}

\begin{figure}[h]
 \includegraphics[width=0.7\linewidth]{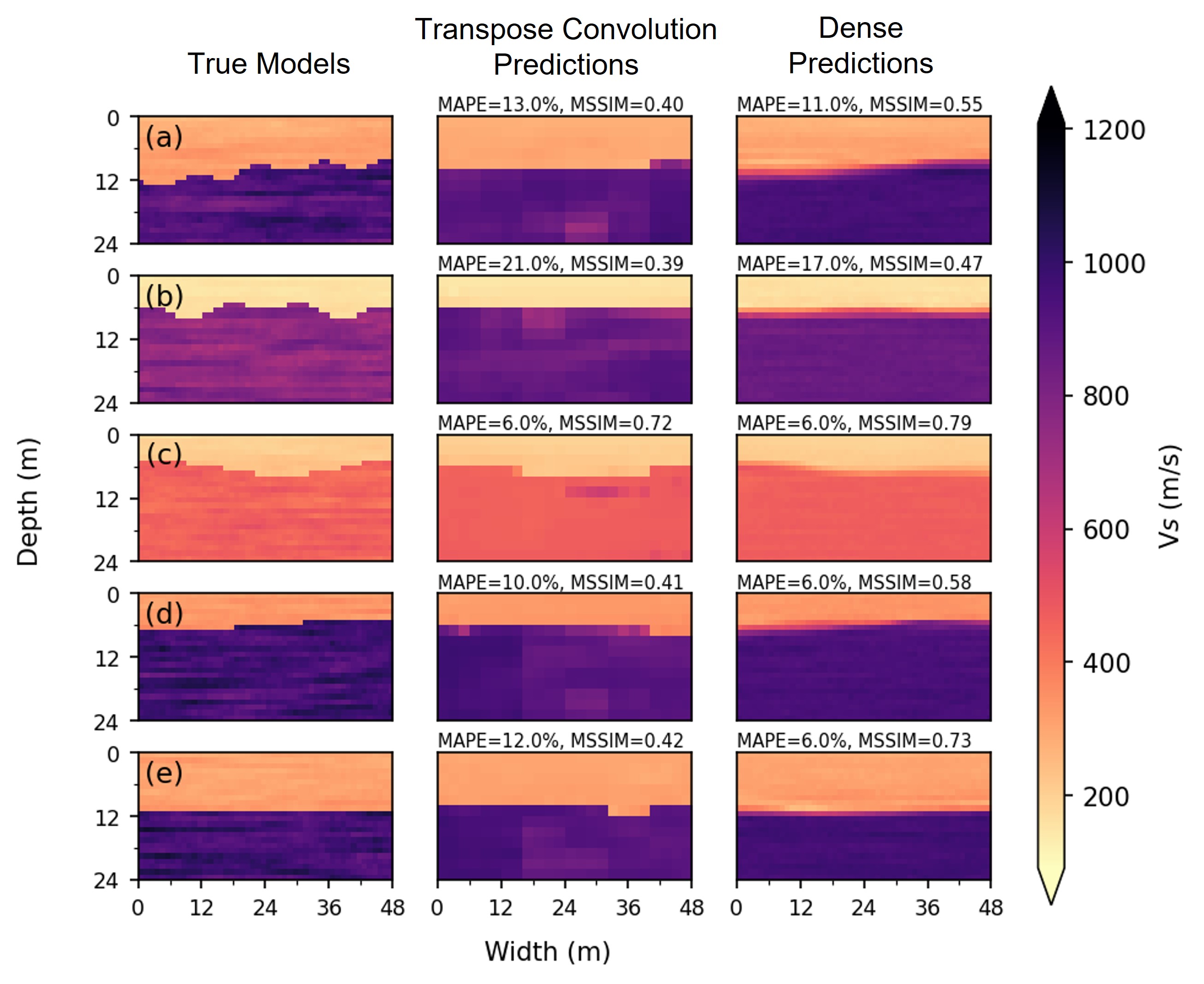}
 \centering
 \vspace{-0.5cm}
   \caption{Comparisons between five randomly selected subsurface shear-wave velocity (Vs) images from the testing dataset (left column) and the predictions made by the Transpose Convolution convolutional neural network (CNN, middle column) and the Dense CNN (right column). Each row has been labeled (a-e) for ease of discussion. The mean absolute percent error (MAPE) and mean structural similarity (MSSIM) index of each predicted image is listed above the image. Note that lower values of MAPE correspond to better Vs value predictions, while higher values of MSSIM correspond to better layer boundary predictions.}
   \label{fig:decoder_pred}
\end{figure}

We now review the predictive accuracy of both networks.~\cref{fig:decoder_pred} presents a set of randomly selected subsurface Vs images from the testing dataset (left column), the Transpose Convolution network’s predictions for these images (middle column), and the Dense network’s predictions for these images (right column). Note that while these images were randomly selected, they have been plotted in order of decreasing layer boundary complexity for ease of discussion and visualization.

Overall, across all 20,000 testing models, the Transpose Convolution network has an MAPE of 11.3\% and MSSIM index of 0.54. Recall that lower values of MAPE correspond to more accurate Vs predictions, while higher values of MSSIM correspond to more accurate layer boundary predictions.  Given the complexity of predicting Vs subsurface profiles from dispersion images, an MAPE of 11.3\% is considered adequate and only a few percentage points higher than the best MAPEs predicted by \citet{abb} for their frequency-velocity CNN. Additionally, a lower MSSIM index is expected; because Transpose Convolution layers followed by UpSampling layers typically cause pixelation in predicted images, the boundary between soil layers is poorly predicted, causing a lower MSSIM index (compare this network’s prediction to the true image in row c of~\cref{fig:decoder_pred}). 

In comparison, we find that the performance of the Dense CNN is superior to the Transpose Convolution CNN, as it has an overall lower MAPE of 8.5\% and a higher MSSIM index of 0.65. This network likewise predicts Vs values of the subsurface relatively well but performs poorly when predicting some of the more complex layer boundaries. For example, compare the true image in row b of~\cref{fig:decoder_pred} to the Dense CNN’s prediction. The highly undulating layer boundary in the true image is not captured by the network’s prediction and is instead predicted as a horizontal line. This smoothing of complex layer boundaries is somewhat expected because convolutional layers inherently blur edge and boundary features, thus resulting in smoother outputs. Nonetheless, the Dense network performs reasonably well in predicting subsurface Vs values and less complex layer boundaries and should be selected over the Transpose Convolution network due to its higher predictive accuracy. However, this network selection ignores the internal workings of the CNNs, meaning the selection of the CNN is made without understanding the explainability of the network.

\begin{figure}
 \includegraphics[width=0.6\linewidth]{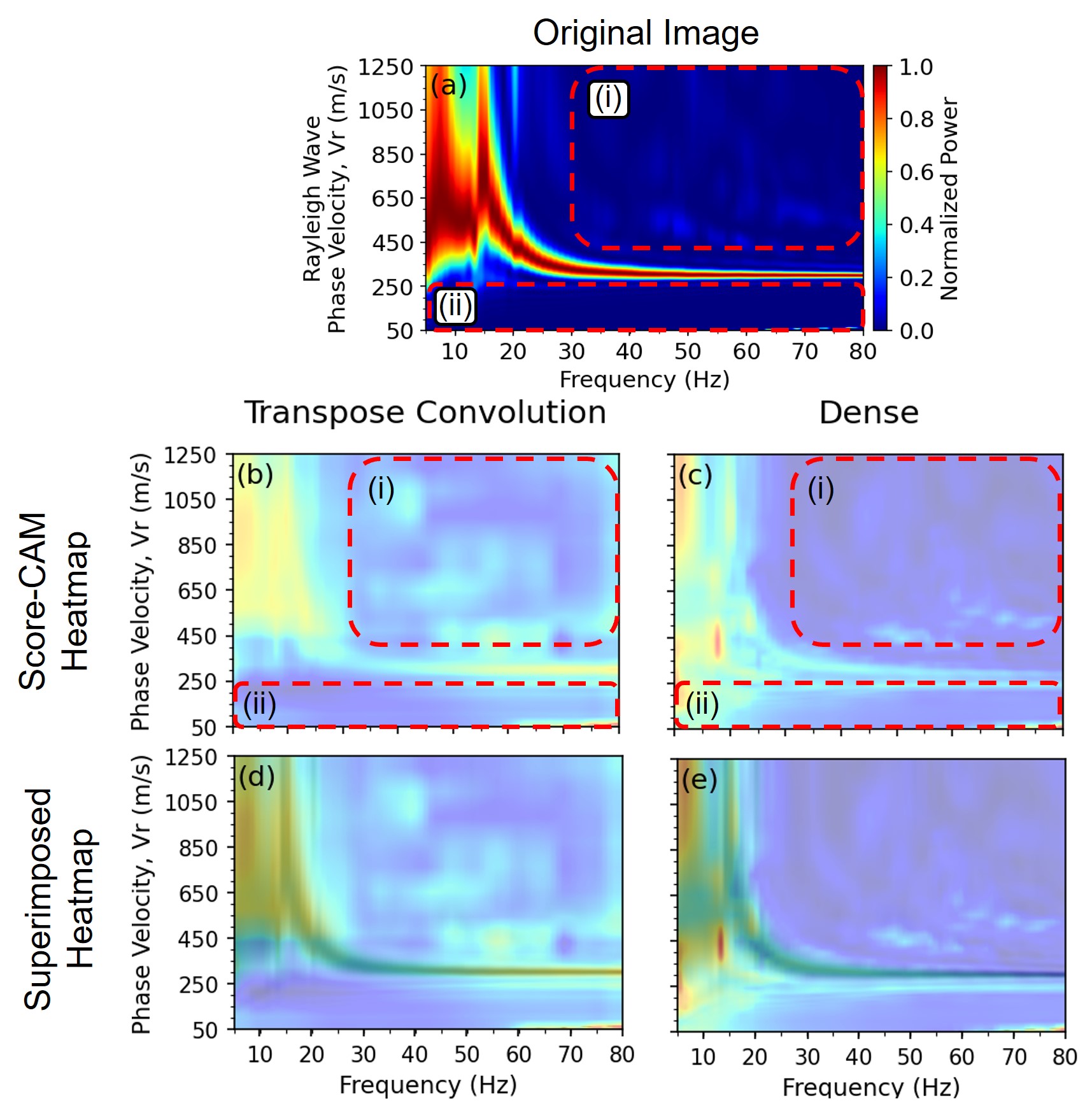}
 \centering
 \vspace{-0.4cm}
   \caption{(a) A randomly selected dispersion input image from the training dataset, the average heatmaps for all layers in the (b) Transpose Convolution network and the (c) Dense network generated using Score-CAM with respect to the selected dispersion input image, and (d, e) the respective heatmaps superimposed on the original input image (shown in translucent grayscale for clarity). Note that warmer colors indicate regions that positively influence the CNN’s decisions, while cooler colors indicate regions that do not influence the CNN’s decisions. Regions containing less important information in the dispersion image (a-i, a-ii) are outlined in red and are shown again in the associated heatmaps.}
   \label{fig:decoder_score}
   \vspace{0.2cm}
 \includegraphics[width=0.8\linewidth]{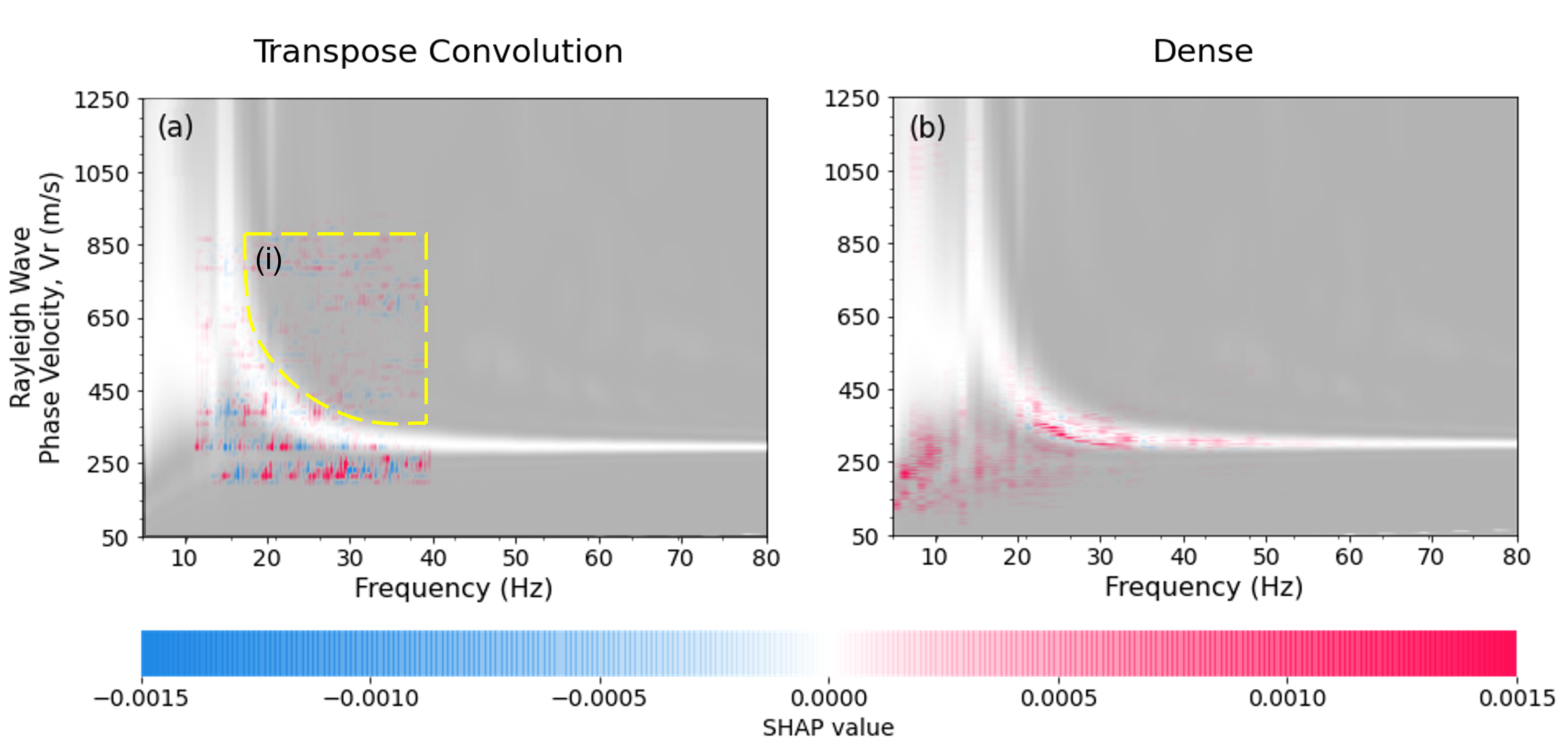}
 \centering
 \vspace{-0.4cm}
   \caption{The SHAP values calculated using Deep SHAP for the selected input image shown as~\cref{fig:decoder_score}a for the (a) Transpose Convolution network and (b) Dense network. Red colors indicate regions that positively influence the CNN’s decisions, blue colors indicate regions that do not influence the CNN’s decisions, and the magnitudes of the SHAP values indicate the significance of the influence.}
   \label{fig:decoder_shap}
\end{figure}

Recall that in the input dispersion image, the high-power modal dispersion trend provides a significant amount of information about the subsurface. Therefore, we hypothesize that the network must learn the most from this high-power modal trend (i.e., regions with the highest normalized power), as it provides physical information about the subsurface. To assess this hypothesis, we evaluate the descriptive accuracy of the networks using the Score-CAM algorithm to generate heatmaps that show where the CNN is “looking” when it learns from the training dataset. These heatmaps highlight regions of the input dispersion image that are most influential to the predictions. To illustrate the results of Score-CAM for both networks, an input dispersion image was randomly selected as an example (\cref{fig:decoder_score}a). Figs.~\labelcref{fig:decoder_score}b and c show the heatmaps generated using Score-CAM for this image for the Transpose Convolution and Dense networks, respectively, and these heatmaps are superimposed onto a translucent grayscale version of the input dispersion image in Figs.~\labelcref{fig:decoder_score}d and e. Regions of the dispersion image that should theoretically contain less important information (i.e., low-power non-modal regions, highlighted as i and ii) are outlined in red in~\cref{fig:decoder_score}a and these regions are shown again in the heatmaps in Figs.~\labelcref{fig:decoder_score}b and c. 

The warmer colors in Score-CAM heatmaps indicate regions of positive influence on the CNN’s decisions, while cooler colors indicate regions of no influence. In Figs.~\labelcref{fig:decoder_score}b and c, the portion of the dispersion image containing high normalized power is clearly visible as a warmer region, meaning both networks strongly consider this part of the input image when learning to make predictions. Additionally, a large region of the Transpose Convolution network's heatmap (\cref{fig:decoder_score}b-i) is highlighted as a warm region, which indicates that this region affects its predictions. This explanation of the prediction is non-physical, as the highlighted region corresponds to dispersion data not associated with the high-power modal trend and not used during typical surface wave inversions (\cref{fig:decoder_score}a-i). Conversely, very little of this region is highlighted in the Dense network's heatmap (\cref{fig:decoder_score}c-i), indicating that it does not consider this portion of the input image when making predictions. Additionally, although the bottom region of the original dispersion image (\cref{fig:decoder_score}a-ii) contains no significant information, the Transpose Convolution network strongly considers the bottom right corner of the image to be important (\cref{fig:decoder_score}b-ii), while the Dense network considers regions in the bottom left and right corners as significant (\cref{fig:decoder_score}c-ii). These results are typical for many other input images reviewed throughout this study, although we present only one image in~\cref{fig:decoder_score} for brevity.

We now use another method, Deep SHAP, to provide another perspective on the CNN’s logic. Using the same input dispersion image as~\cref{fig:decoder_score}a, the SHAP values measuring the importance of regional differences relative to the first layer in the network are shown in Figs.~\labelcref{fig:decoder_shap}a and b for the Transpose Convolution and Dense networks, respectively. Note that in~\cref{fig:decoder_shap}, positive SHAP values are red in color, while negative SHAP values are blue in color. The SHAP values reveal that both networks strongly consider portions of the high-power modal trend at low frequencies relative to the rest of the input when making predictions. However, the mix of negative and positive SHAP values indicates the Transpose Convolution network has difficulties identifying the important features in this region.~\cref{fig:decoder_shap}i further shows that the Transpose Convolution network incorrectly considers unimportant parts of the input image (recall~\cref{fig:decoder_score}a-i) as important to its decision making. In contrast, the Dense network considers more of the high-power modal trend as a positive influence on its decision making, despite indicating some of the lower left portions of the dispersion image as erroneously important. Therefore, we find that the Dense network follows our hypothesis more closely than the Transpose Convolution network.

When reviewing the predictive accuracy of the Transpose Convolution and Dense networks, it seems that the Dense network is superior, as it has a higher overall MAPE and MSSIM index. Additionally, the descriptive accuracy of the Dense network is relatively higher than the Transpose Convolution network, although its logic could likely be improved through hyperparameter tuning. Therefore, we choose a Dense layer as the decoder structure for the CNN given its relatively higher predictive and descriptive accuracy, and we perform additional hyperparameter tuning to further improve the network’s descriptive accuracy. We next focus on tuning the kernel size of the convolutional layers used in the network’s encoder structure.

\subsection{Selecting a convolution kernel size in the encoder}

\begin{figure}[t!]
    \centering
 \includegraphics[width=0.7\linewidth]{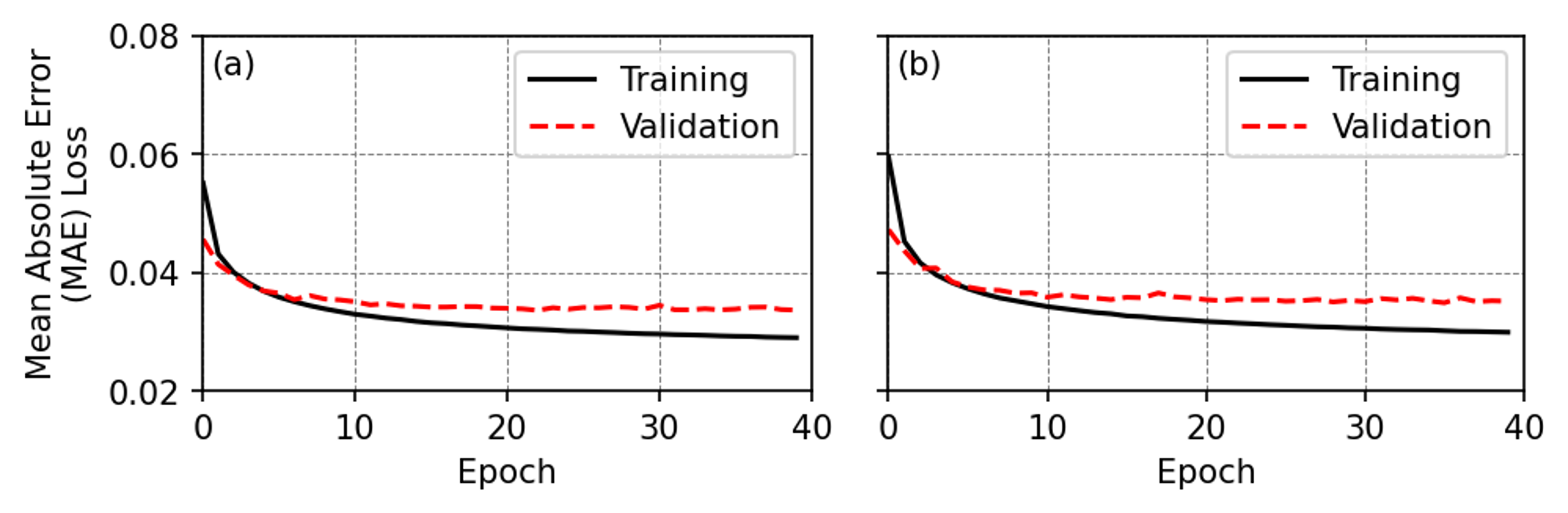}
 \centering
 \vspace{-0.5cm}
   \caption{Mean absolute error (MAE) loss reduction observed for the training and validation datasets during convolutional neural network (CNN) optimization for (a) the 3$\times$kernel network and (b) the 3$\times$kernel network. Note that neither network is overfit.}
   \label{fig:loss}
\end{figure}

We next evaluate two different kernel sizes for the encoder in the Dense network: (1) a traditional square 3$\times$3 kernel for all convolutional layers and (2) an atypical rectangular 3$\times$1 kernel for the first three convolutional layers. We compare the performance of the two kernel sizes in preserving the frequency-dependent information (i.e., velocity dispersion occurring at each frequency) before convolution with other frequencies. To compare kernel sizes, we use the previous Dense network as the baseline (now referred to as the 3$\times$3 network) and compare this to a similar network in which only the kernel size of the first three convolutional layers is changed to 3$\times$1 (referred to as the 3$\times$1 network). Both networks are trained and tested using the same datasets described previously. We use the same learning rate, batch size, number of training epochs, optimizer, and loss function as before. Table~\labelcref{table:deep} provides a detailed description of the 3$\times$3 and 3$\times$1 CNNs, and the kernel sizes of the first three convolutional layers are highlighted to emphasize the difference between the networks. Figs.~\labelcref{fig:loss}a and b show the MAE loss per epoch during training for the 3$\times$3 network and the 3$\times$1 network, respectively. Both networks performed comparably in terms of MAE loss reduction during validation, and we emphasize that neither network shows signs of overfit.

\begin{table}[h!]
\centering
 \caption{A detailed description of the convolutional neural networks (CNNs) designed with 3$\times$3 kernels (left) and 3$\times$1 kernels (right). Note that the only difference between the networks is the kernel size of the first three convolutional layers, which is highlighted.}
 \vspace{0.3cm}
\small
\label{table:deep}
\def\arraystretch{1.2}
\begin{tabular}{>{\centering\arraybackslash}m{3.0cm} >{\centering\arraybackslash}m{1.0cm}| >{\centering\arraybackslash}m{2.4cm}
>{\centering\arraybackslash}m{2.2cm}| >{\centering\arraybackslash}m{2.4cm} >{\centering\arraybackslash}m{2.2cm}}
 \toprule
 \multicolumn{2}{c}{} &
 \multicolumn{2}{c}{3$\times$3 Network} & \multicolumn{2}{c}{3$\times$1 Network}\\
 \midrule
 Network Layer Type	& Filters & Kernel/Pool Size & Output Shape & Kernel/Pool Size & Output Shape\\
 \midrule
 Input & & & 400$\times$76$\times$1 (Dispersion Image) & & 	400$\times$76$\times$1 (Dispersion Image)\\
 2D Convolution & 32 & \cellcolor{mygray}(3, 3) & 398$\times$74$\times$32 & \cellcolor{mygray}(3, 1) & 398$\times$76$\times$32\\
 2D Max Pooling & & (3, 1) & 132$\times$74$\times$32 & (3, 1) & 132$\times$76$\times$32\\
 2D Convolution & 32 & \cellcolor{mygray}(3, 3) & 130$\times$72$\times$32 & \cellcolor{mygray}(3, 1) & 130$\times$76$\times$32\\
 2D Max Pooling & & (3, 1) & 43$\times$72$\times$32 & (3, 1) & 43$\times$76$\times$32\\
 2D Convolution & 64 & \cellcolor{mygray}(3, 3) & 41$\times$70$\times$64 & \cellcolor{mygray}(3, 1) & 41$\times$76$\times$64\\
 2D Max Pooling & & (1, 3) & 41$\times$23$\times$64 & (1, 3) & 41$\times$25$\times$64\\
 2D Convolution & 128 & (3, 3) & 39$\times$21$\times$128 & (3, 3) & 39$\times$23$\times$128\\
 2D Max Pooling & & (3, 3) & 13$\times$7$\times$128 & (3, 3) & 13$\times$7$\times$128\\
 2D Convolution & 128 & (3, 3) & 11$\times$5$\times$128 & (3, 3) & 11$\times$5$\times$128\\
 Flatten & & & 7040 & & 7040\\
 Dense & & & 1152 (Flattened Vs Image) & & 1152 (Flattened Vs Image)\\
 \bottomrule
\end{tabular}
\end{table}

\begin{figure}[hbt!]
 \includegraphics[width=0.7\linewidth]{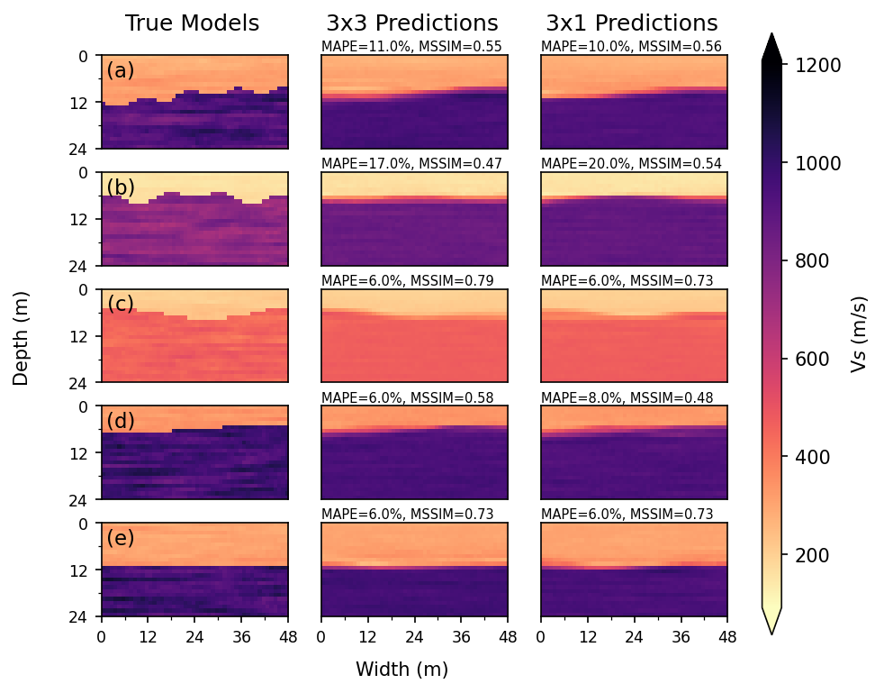}
 \centering
 \vspace{-0.5cm}
   \caption{Comparisons between five randomly selected subsurface shear-wave velocity (Vs) images from the testing dataset (left column) and the predictions made by the 3$\times$3 convolutional neural network (CNN, middle column) and the 3$\times$1 CNN (right column). Each row has been labeled (a-e) for ease of discussion. The mean absolute percent error (MAPE) and mean structural similarity (MSSIM) index of each predicted image is listed above the image. Note that lower values of MAPE correspond to better Vs value predictions, while higher values of MSSIM correspond to better layer boundary predictions.}
   \label{fig:deep_pred}
\end{figure}

We now compare the predictions of both networks for the same randomly selected set of images from the testing dataset shown in~\cref{fig:decoder_pred} for consistency. These true subsurface images are shown again as the left column in~\cref{fig:deep_pred}, while the corresponding predictions of the 3$\times$3 and 3$\times$1 networks are shown as the middle and right columns, respectively. In~\cref{fig:deep_pred}, each prediction made by the 3$\times$3 network (middle column) is visually similar to each respective prediction made by the 3$\times$1 network (right column), and the MAPE and MSSIM index values calculated for these predictions are nearly identical. Overall, the 3$\times$3 network has an MAPE of 8.5\% and MSSIM index of 0.65, and the 3$\times$1 network has similar values of 8.8\% and 0.65, respectively. These networks predict Vs values of the subsurface quite well, but perform poorly when predicting the more complex layer boundaries. For example, compare the true image in row b of~\cref{fig:deep_pred} to each network’s prediction. The highly undulating layer boundary in the true image is not captured by the predictions of either network and is instead predicted as a horizontal line. This is again expected because convolutional layers inherently blur edge and boundary features as the network learns, thus resulting in smoother outputs. Despite this blurring of complex layer boundaries, the predictions capture the average depth of the boundaries quite well.   

Because both networks perform similarly, the challenge is now choosing between the 3$\times$3 network and the 3$\times$1 network (i.e., which kernel size to use in the first three convolutional layers). The simplest choice is to select the 3$\times$3 kernel, as it provides the marginally lower MAPE and the same MSSIM index. However, this choice would again ignore the descriptive accuracy of each CNN, and so we propose using post hoc explainers to provide more insight regarding the choice of our final network.

\begin{figure}[hbt!]
 \includegraphics[width=0.6\linewidth]{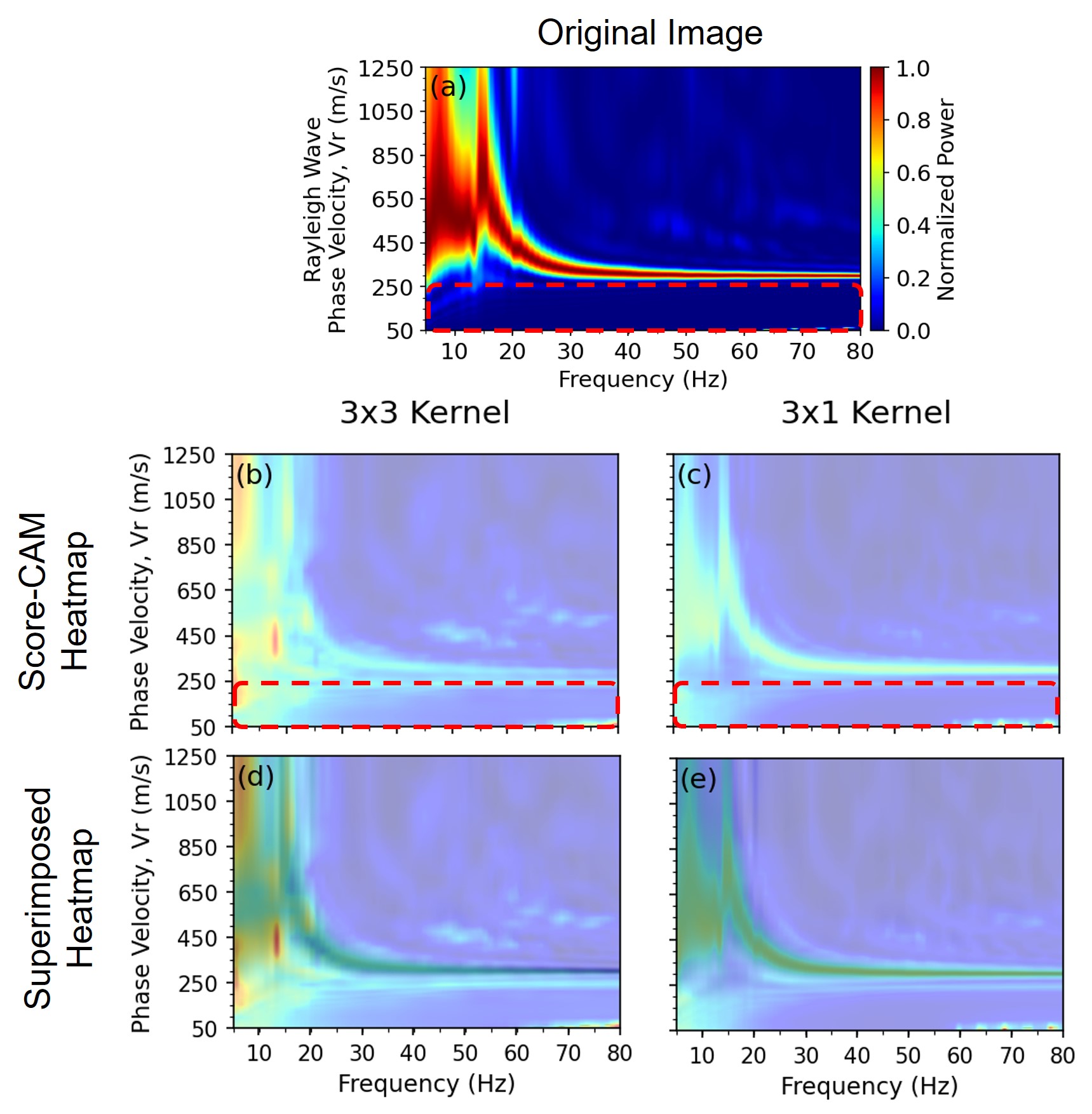}
 \centering
 \vspace{-0.4cm}
   \caption{(a) A randomly selected dispersion input image from the training dataset, the average heatmaps for all layers in the (b) 3$\times$3 network and the (c) 3$\times$1 network generated using Score-CAM with respect to the selected dispersion input image, and (d, e) the respective heatmaps superimposed on the original input image (shown in translucent grayscale for clarity). Note that warmer colors indicate regions that positively influence the CNN’s decisions, while cooler colors indicate regions that do not influence the CNN’s decisions. The region outlined in red in (a) illustrates a portion of the input image containing less important information.}
   \label{fig:deep_score}
\end{figure}

We again evaluate each network’s descriptive accuracy by developing heatmaps that provide insight into what the network considers important in each dispersion input image. We require that the network learns the most information from the high-power modal trend in the input image. The results of Score-CAM averaged across all layers in either network are shown in~\cref{fig:deep_score}, where the same input dispersion image shown in~\cref{fig:decoder_score}a is shown again as~\cref{fig:deep_score}a, the heatmaps generated using Score-CAM for this image are shown as Figs.~\labelcref{fig:deep_score}b and c for the 3$\times$3 and 3$\times$1 networks, respectively, and these heatmaps are superimposed onto a translucent grayscale version of the input dispersion image in Figs.~\labelcref{fig:deep_score}d and e. We also used Deep SHAP to quantify the relative importance of each part of the input image and found that the Deep SHAP results confirmed the trends present in the Score-CAM heatmaps as before. Therefore, we omit the Deep SHAP plots for brevity. 

Although both networks seem to perform similarly when considering their predictions, MAPEs, and MSSIM indices, the results of Score-CAM show that the networks’ learned features differ. In~\cref{fig:deep_score}b, we see that while the 3$\times$3 network’s heatmap highlights the general shape of the high-power modal trend in the input image, it also strongly considers the entire left portion of the image.~\cref{fig:deep_score}c shows that the 3$\times$1 network can better distinguish the general shape of the high-power modal trend from surrounding features. Despite this, both networks still fixate on some unimportant features, such as regions in the bottom left and right corners of the input image where only low-power dispersion data are present in the input image, as illustrated by the red outlined region in~\cref{fig:deep_score}a. Note that while we present a single input image in~\cref{fig:deep_score}, this finding is typical across other input images analyzed in this study. 

\begin{figure}
 \includegraphics[width=0.8\linewidth]{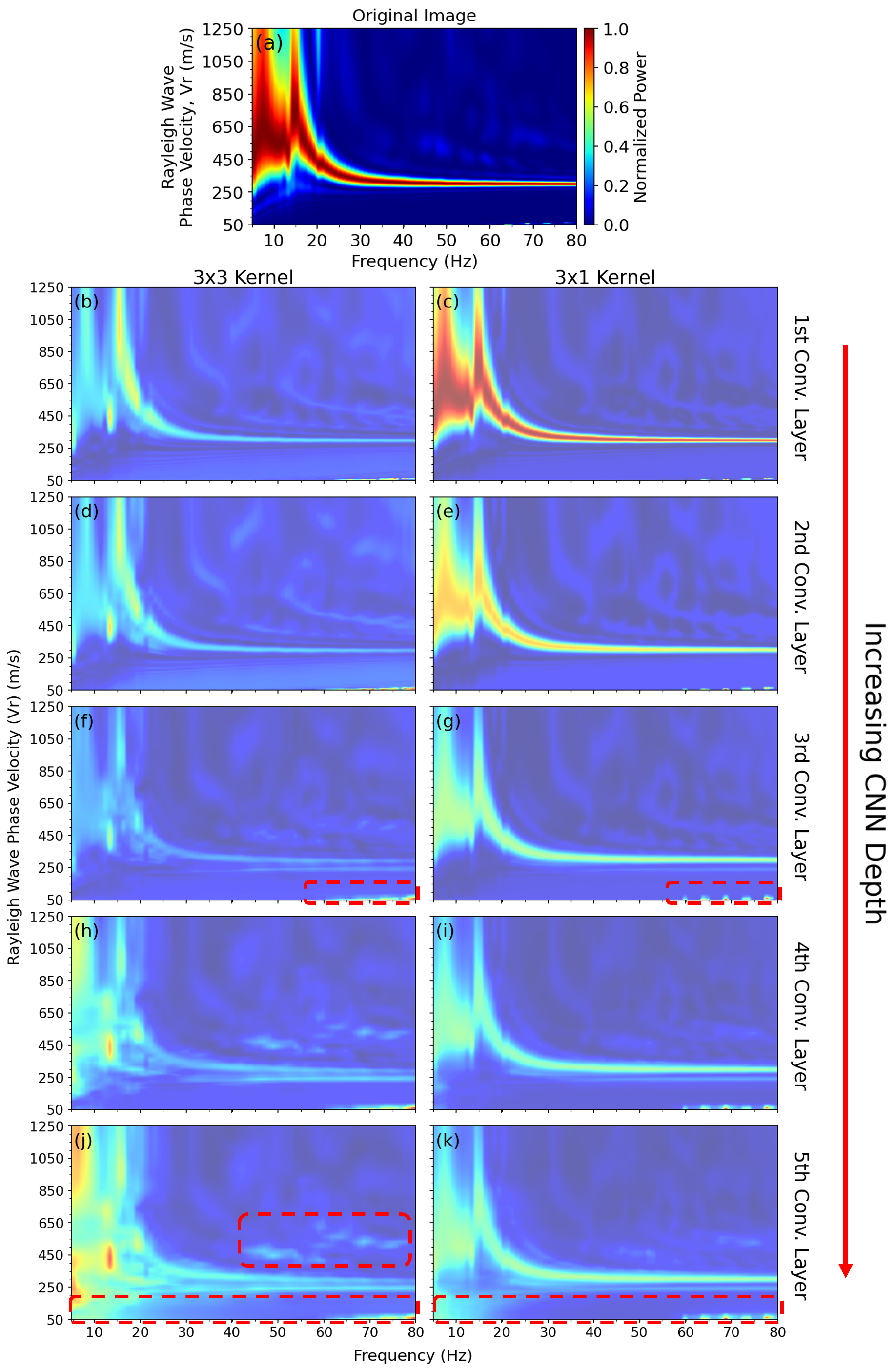}
 \centering
 \vspace{-0.5cm}
   \caption{(a) A selected dispersion input image from the training dataset and the average Score-CAM heatmaps for each convolutional layer in the 3$\times$3 network (left column) and the 3$\times$1 network (right column). Note that warmer colors indicate regions that positively influence the CNN’s decisions, while cooler colors indicate regions that do not influence the CNN’s decisions. Increasing row depth corresponds to increasing CNN depth, as shown to the right of the figure.}
   \label{fig:scorelong}
\end{figure}

Both explainability methods reveal a fixation on regions that are typically unimportant in dispersion images used for standard surface wave testing, and we believe both networks are excessively complex, leading to learning from noise or insignificant features. Therefore, we now use Score-CAM to develop averaged heatmaps for each convolutional layer in both networks to visualize when the network begins to learn from these insignificant features. To do so, we first calculate a heatmap for each individual layer in the network, then simply average the results at each convolutional layer. The results are shown in~\cref{fig:scorelong}, where the original input image is repeated as~\cref{fig:scorelong}a and the heatmaps for the 3$\times$3 network and 3$\times$1 network are shown in the left and right columns, respectively. Each convolutional layer is labeled to the right of each row and increasing row depth corresponds to increasing network depth. Note that these heatmaps are shown at a higher intensity and are not superimposed on the original input image for better visualization. 

In~\cref{fig:scorelong}, we see that both networks learn a significant amount of information regarding the high-power modal trend in the input image within the first two convolutional layers (Figs.~\labelcref{fig:scorelong}b-e). However, beginning at the third convolutional layer, we see the networks begin to focus on noise in the bottom right corner (outlined in red). At the fourth convolutional layer, the 3$\times$3 network begins focusing on additional noisy features at the bottom left corner. By the final convolutional layer, both networks focus on noisy features (outlined in red), with the 3$\times$3 network considering relatively more noisy features as relevant to its decision making. We believe this fixation on low-power noise indicates a type of network overfit, although neither network shows signs of conventional overfit when considering their predictive accuracies (recall~\cref{fig:loss}). While the 3$\times$1 network has a relatively higher descriptive accuracy compared to the 3$\times$3 network, we find that both networks begin capturing noisy features, and thus we create shallower architectures as a means to improve their descriptive accuracy.

\subsection{Selecting the network depth}

As the CNN learns most of the high-power modal trend features in the first two layers, we shorten the previous two architectures to have only two 2D convolutional layers instead of five. Additionally, our only pooling layer now uses a 3$\times$3 pooling area instead of a 3$\times$1 area, as this dramatically reduces the number of trainable parameters from approximately 350 million to 100 million, thus reducing the training time from approximately 26 hours to 10 hours. We retain the final Dense layer as our decoder structure given its superior performance compared to the decoder structure of the Transpose Convolution CNN. We also use the same learning rate, batch size, number of training epochs, optimizer, and loss function as the previous networks. A detailed description of both the shallow 3$\times$3 CNN and the shallow 3$\times$1 CNN are provided in Table~\labelcref{table:shallow}. We note that the loss reduction during training and validation showed that neither network overfits.

\begin{table}
\caption{A detailed description of the shallow convolutional neural networks (CNNs) designed with 3$\times$3 kernels (left) and 3$\times$1 kernels (right).}
\label{table:shallow}
\vspace{0.3cm}
\small
\centering
\def\arraystretch{1.2}
\begin{tabular}{>{\centering\arraybackslash}m{2.6cm} >{\centering\arraybackslash}m{1.3cm}| >{\centering\arraybackslash}m{1.6cm}
>{\centering\arraybackslash}m{2.8cm}| >{\centering\arraybackslash}m{1.8cm} >{\centering\arraybackslash}m{2.8cm}}
 \toprule
 \multicolumn{2}{c}{} &
 \multicolumn{2}{c}{Shallow 3$\times$3 Network} & \multicolumn{2}{c}{Shallow 3$\times$1 Network}\\
 \midrule
 Network Layer Type	& Filters & Kernel/Pool Size & Output Shape & Kernel/Pool Size & Output Shape\\
 \midrule
 Input & & & 400$\times$76$\times$1 (Dispersion Image) & & 	400$\times$76$\times$1 (Dispersion Image)\\
 2D Convolution & 32 & (3, 3) & 398$\times$74$\times$32 & (3, 1) & 398$\times$76$\times$32\\
 2D Max Pooling & & (3, 3) & 132$\times$42$\times$32 & (3, 3) & 132$\times$25$\times$32\\
 2D Convolution & 32 & (3, 3) & 130$\times$22$\times$32 & (3, 1) & 130$\times$25$\times$32\\
 Flatten & & & 91520 & & 104000\\
 Dense & & & 1152 (Flattened Vs Image) & & 1152 (Flattened Vs Image)\\
 \bottomrule
\end{tabular}
\end{table}

\begin{table}[h]
\begin{center}
\caption{A comparison between the MAPE (shown as a percentage) and MSSIM index (shown in parentheses) between the deep and shallow 3$\times$3 and 3$\times$1 networks.}
\label{table:accuracy}
\vspace{0.3cm}
\small
\def\arraystretch{1}
\begin{tabular}{>{\centering\arraybackslash}m{3.8cm} >{\centering\arraybackslash}m{2.2cm} >{\centering\arraybackslash}m{2.2cm}}
 \toprule
 & 3$\times$3 Network & 3$\times$1 Network\\
 \midrule
 Deep\\ (5 convolutional layers) & 8.5\% (0.65) & 8.8\% (0.65)\\
 Shallow\\ (2 convolutional layers) & 11.2\% (0.58) & 12.1\% (0.57)\\
 \bottomrule
\end{tabular}
\end{center}
\end{table}

\begin{figure}[h]
 \includegraphics[width=0.7\linewidth]{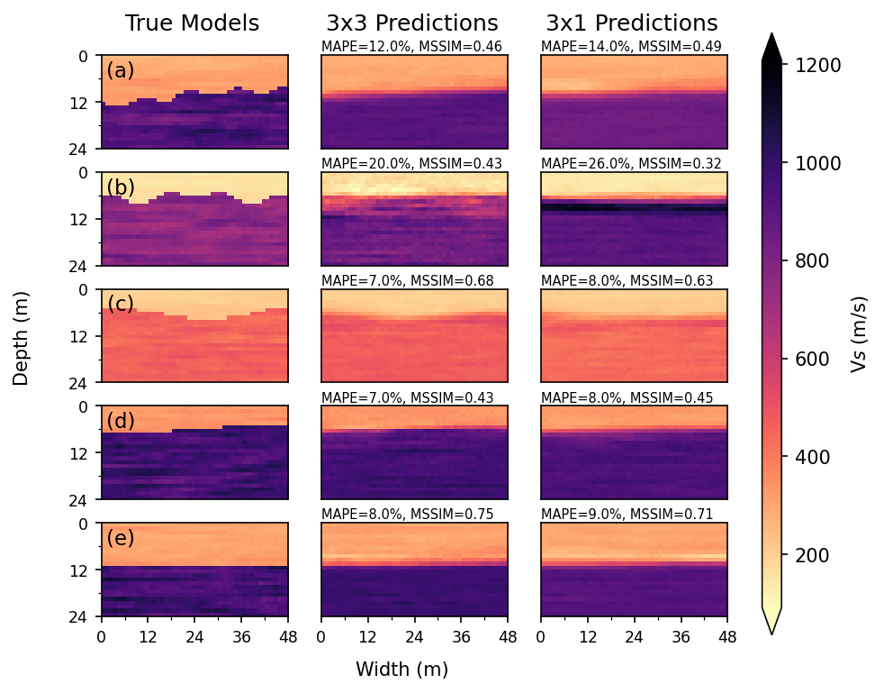}
 \centering
 \vspace{-0.4cm}
   \caption{Comparisons between five randomly selected subsurface shear-wave velocity (Vs) images from the testing dataset (left column) and the predictions made by the shallow 3$\times$3 convolutional neural network (CNN, middle column) and the shallow 3$\times$1 CNN (right column). Each row has been labeled (a-e) for ease of discussion. The mean absolute percent error (MAPE) and mean structural similarity (MSSIM) index of each predicted image is listed above the image. Note that lower values of MAPE correspond to better Vs value predictions, while higher values of MSSIM correspond to better layer boundary predictions.}
   \label{fig:shallow_pred}
\end{figure}

We now compare the predictions of each revised, shallow network. The same subset of images selected from the testing dataset shown in the left column in~\cref{fig:decoder_pred} is presented again as the left column in~\cref{fig:shallow_pred}. The predictions of these images using the shallow 3$\times$3 network and the shallow 3$\times$1 network are shown as the middle and right columns, respectively, in~\cref{fig:shallow_pred}. When reviewing the results shown in~\cref{fig:shallow_pred}, we find the shallow networks follow similar trends to the deeper networks shown previously. Specifically, we find that both shallow networks have: (1) relatively low MAPE values, meaning Vs values are generally well-predicted; (2) relatively low MSSIM values for more complex models, which correspond to poor layer boundary predictions; and (3) smoothed layer boundaries resulting from the convolutional layers in the network. Overall, the shallow 3$\times$3 network has an MAPE of 11.2\% and an MSSIM index of 0.58, while the shallow 3$\times$1 network has an MAPE of 12.1\% and an MSSIM index of 0.57. In general, the networks provide similar predictions to one another, although the shallow 3$\times$1 network typically has slightly higher errors. Despite their differences, both shallow networks perform poorly relative to the deeper networks shown in the previous section, as shown in Table~\labelcref{table:accuracy}. Recall, however, that the choice to use a shallow network was based on the network’s fixation on unimportant features seen in the explanations of the deeper CNNs (\cref{fig:scorelong}). Therefore, we attempt to explain the predictions of the shallow networks.

\begin{figure}[hbt!]
 \includegraphics[width=0.6\linewidth]{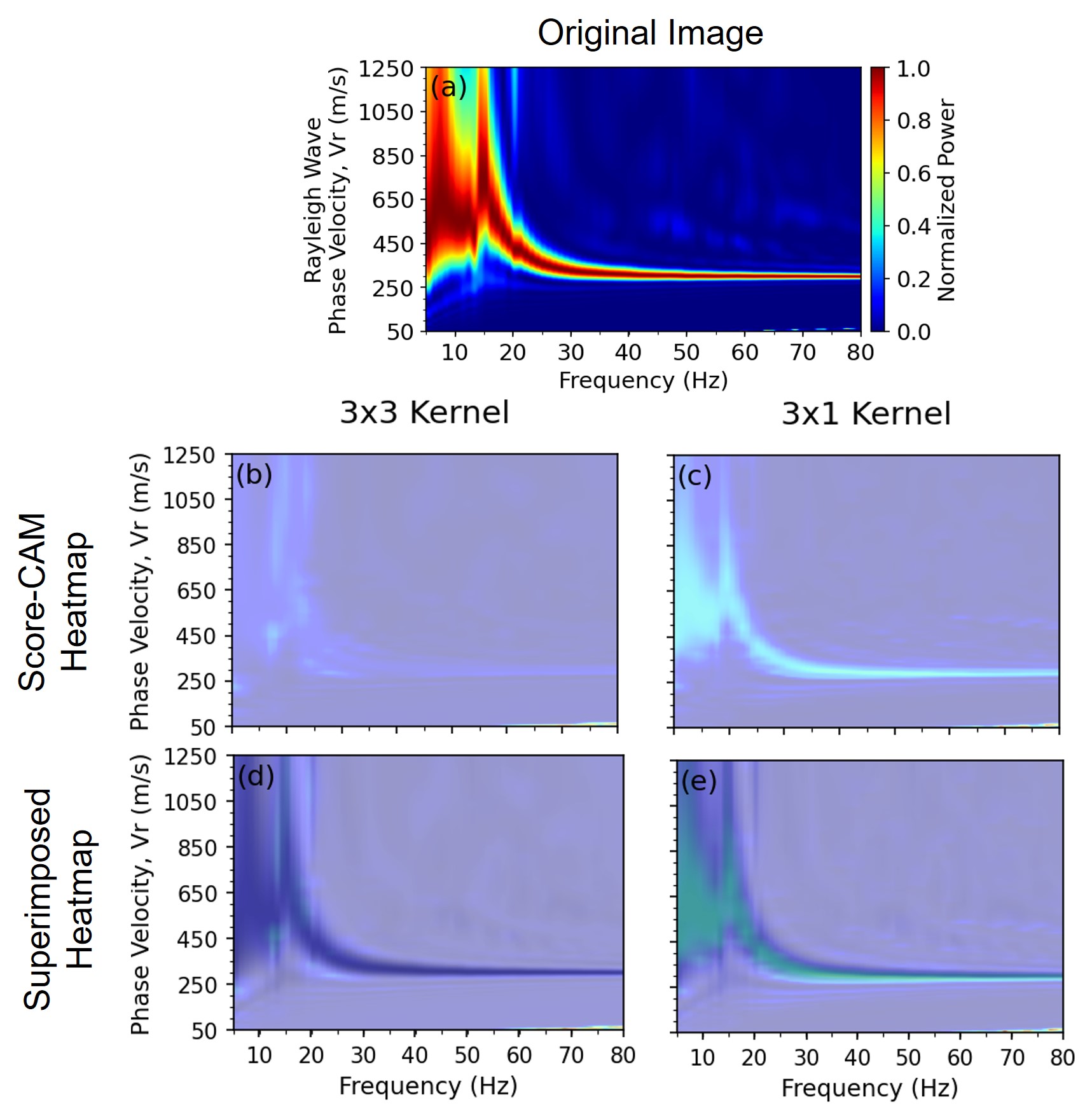}
 \centering
 \vspace{-0.4cm}
   \caption{(a) A selected dispersion input image from the training dataset, the average heatmaps for all layers in the (b) shallow 3$\times$3 kernel network and the (c) shallow 3$\times$1 kernel network generated using Score-CAM with respect to the selected dispersion input image, and (d, e) the respective heatmaps superimposed on the original input image (shown in translucent grayscale for clarity). Warmer colors indicate regions that positively influence the CNN’s decisions, while cooler colors indicate regions that do not influence the CNN’s decisions.}
   \label{fig:shallow_score}
\end{figure}

To evaluate the descriptive accuracy of our networks, we again use Score-CAM to explain our networks’ decisions. The results of Score-CAM averaged across all layers in either network are shown in~\cref{fig:shallow_score} for the same input dispersion image in~\cref{fig:decoder_score}a. The heatmaps generated using Score-CAM for this image are shown as Figs.~\labelcref{fig:shallow_score}b and c for the shallow 3$\times$3 and 3$\times$1 networks, respectively, and these heatmaps are superimposed onto a translucent grayscale version of the input image in Figs.~\labelcref{fig:shallow_score}d and e.

The Score-CAM results shown in Figs.~\labelcref{fig:shallow_score}b and d reveal that the shallow 3$\times$3 network has difficulty distinguishing features in the input image, while Figs.~\labelcref{fig:shallow_score}c and e show the shallow 3$\times$1 network considers nearly the entire high-power modal trend as important to its decision making. Further, unlike the deep 3$\times$1 network in~\cref{fig:deep_score}e, the shallow 3$\times$1 network does not fixate on other low-power features present in the input dispersion image (\cref{fig:shallow_score}e), which we believe indicates that this network is not excessively complex. As with the previous hyperparameter tuning, these methods were used to analyze a variety of randomly selected input dispersion images, and the trends presented here are representative of the training dataset. Although the shallow 3$\times$1 network has a relatively low predictive accuracy compared to other dense-layer networks considered in this study (recall~\cref{table:accuracy}), we find that it has the highest descriptive accuracy and therefore makes the most logical predictions. Therefore, we recommend the CNN with two convolutional layers using a 3$\times$1 kernel as our final network, as it yields a high descriptive accuracy without a considerable loss to predictive accuracy. 

We recommend a shallow two-layer CNN that yields high descriptive accuracy but a relatively lower predictive accuracy, as we emphasize creating a physics-aware network that makes logical predictions. \hl{Recall that we create this "physics-aware" network by guiding it towards learning information about the physical properties of the subsurface represented by the high-power modal trend in each dispersion image.} Although black box models such as CNNs are typically designed to have high predictive accuracy, they may sometimes learn incorrect physical behaviors, which may not be noticeable when tested on the local dataset, i.e., the network is interpolating \citep{kumar}. This leads to issues when trying to generalize the model to conditions absent from the local dataset (i.e., different subsurface conditions, type of active source, etc.). By focusing on improving our network’s descriptive accuracy, we create a more robust model with respect to modeling the physics behind our problem, thus leading to more trustworthy predictions. However, we note that shallower networks are typically sensitive to noisy inputs (recall that deeper layers in CNNs capture more complex information in the input image). Therefore, while we recommend beginning with the shallow 3$\times$1 network proposed herein, we further recommend that researchers looking to develop networks for field applications should consider adding additional denoising layers to the model. Alternatively, researchers may add noise to their training dataset, resulting in a network that is more robust to noisy field data while still being physics-aware.

\section{Conclusion}

As deep learning models, such as CNNs, have increased in popularity in recent years, the demand for explainable models has grown significantly. Although post hoc explanations have offered insights into the decision-making process of CNNs, the neural network design largely relies on brute-force optimization of the architecture that yields the lowest predictive error. We propose a framework for designing deep neural networks using explainability. We demonstrate that the descriptive accuracy of a network offers insights and guidelines for selecting its hyperparameters. We present the XAI-guided design of a CNN for predicting shallow subsurface 2D Vs profiles based on input dispersion images obtained from active-source wavefields recorded by a linear array of surface sensors. We develop initial CNNs trained on subsurface Vs and surface wave dispersion image pairs. We define the predictive accuracy of each network using two metrics: the Mean Absolute Percentage Error (MAPE), which captures the overall prediction of Vs, and the Mean Structural Similarity Index (MSSIM), which captures the interface boundary. The intial Transpose Convolution and Dense networks show a relatively low overall MAPE and high MSSIM value, indicating good performance. However, post hoc explainability methods, such as Score-CAM and Deep SHAP, show that these networks provide predictions based on some of the low-power regions in the input dispersion images containing no meaningful information. We select the Dense network as a baseline due to its relatively higher descriptive accuracy compared to the Transpose Convolution network, and further explore the predictive and descriptive accuracies of using different kernel sizes: a conventional square 3$\times$3 kernel and a non-conventional rectangular 3$\times$1 kernel. The new 3$\times$3 and 3$\times$1 networks result in relatively similar predictive accuracies. However, post hoc explanations reveal that these networks use regions of low power in the input images for their predictions, while also focusing on non-physical, unimportant, and noisy features. We minimize this effect by altering the network depth to be relatively shallow (i.e., fewer convolutional layers). Although the shallow 3$\times$3  and 3$\times$1 networks perform slightly worse than the previous deeper networks in predictive accuracy, their model explanations showed that the 3$\times$1 network made logical decisions focused primarily on the high-power modal trends and avoided overfit (i.e., had higher descriptive accuracy). \hl{In this work, we select only three hyperparameters to demonstrate the hyperparameter tuning procedure. However, this process can be applied to any number of hyperparameters.} Through this iterative process, we demonstrate the usefulness and necessity of using model explanations to influence the development and design of physics-aware deep learning models. 

\section*{Acknowledgments}
This material is based upon work supported by the National Science Foundation Graduate Research Fellowship under Grant No. DGE – 1610403. However, any opinions, findings, and conclusions or recommendations expressed in this material are those of the authors and do not necessarily reflect the views of the National Science Foundation.

The open-source software DENISE~\citep{kohn,ko} was used to perform all wave propagation simulations conducted for this study. The Texas Advanced Computing Center’s (TACC’s) clusters Stampede2 and Frontera were used during the construction of our dataset and the training and testing of our CNNs, with an allocation provided by the DesignSafe-CyberInfrastructure \citep{rat}. The open-source machine learning libraries TensorFlow \citep{aba} and Keras \citep{chol} were used to develop, train, and test the CNNs presented in this study. The wavefield transformations used to develop the dispersion images were performed using the open-source Python package \emph{swprocess} \citep{vant}.

\section*{Data availability}
Further information regarding the networks and dataset (including subsurface models, wavefields, and dispersion images) used in this study is available from the corresponding author upon reasonable request.

\bibliographystyle{elsarticle-harv}
\bibliography{bibliography.bib}

% \bsp % ``This paper has been produced using the Blackwell
     %   Publishing GJI \LaTeXe\ class file.''

\label{lastpage}

\end{document}